\documentclass[10pt,twocolumn,letterpaper]{article}
\usepackage{amsmath}
\usepackage{amssymb}
\usepackage{times}
\usepackage{color}
\usepackage[dvipsnames]{xcolor}
\usepackage{multicol}
\usepackage{graphicx}
\usepackage{graphics} 
\usepackage{epsfig} 
\usepackage{mathptmx} 
\usepackage{times} 
\usepackage{amsmath} 
\usepackage{amssymb}  
\usepackage{siunitx} 
\usepackage{textcomp}
\usepackage{gensymb} 
\usepackage{booktabs}
\usepackage{multirow}
\usepackage{xcolor}
\let\labelindent\relax
\usepackage{xspace}

\usepackage{makecell}
\usepackage{multicol}
\usepackage{rotating}
\usepackage{courier}
\usepackage{array}
\usepackage{subcaption} 
\usepackage[font=small]{caption}

\usepackage{pifont}
\usepackage[most]{tcolorbox}
\newcommand{\hypbox}[2]{%
\begin{tcolorbox}[colback=white!98!black,colframe=white!30!black,boxsep=1.1pt,top=6.75pt]%
\vspace{1.75pt}%
\textbf{#1}\\[-0.575em]
\noindent\makebox[\textwidth]{\rule{\textwidth}{0.4pt}}
\\[0.25em]
#2
\end{tcolorbox}
}

\usepackage{iccv}              

\usepackage{multirow}

\definecolor{iccvblue}{rgb}{0.21,0.49,0.74}
\usepackage[pagebackref,breaklinks,colorlinks,allcolors=iccvblue]{hyperref}
\usepackage{marvosym}
\def\paperID{4566} 
\def\confName{ICCV}
\def\confYear{2025}

\title{\includegraphics[width=0.12\textwidth]{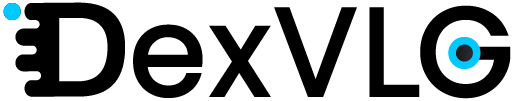}:  \textbf{Dex}terous \textbf{V}ision-\textbf{L}anguage-\textbf{G}rasp Model at Scale}

\author{
  Jiawei He$^{1,2*}$,
  Danshi Li$^{2*}$,
  Xinqiang Yu$^{2*}$,
  Zekun Qi$^{2,3}$,
  Wenyao Zhang$^{2,6,7}$, \\Jiayi Chen$^{2,4}$, Zhaoxiang Zhang$^{5}\textsuperscript{\Letter}$, Zhizheng Zhang$^{2}\textsuperscript{\Letter}$, Li Yi$^{3}\textsuperscript{\Letter}$, He Wang$^{1,2,4}\textsuperscript{\Letter}$
  \\ 
$^1$ BAAI,
$^2$ Galbot,
$^3$ THU,
$^4$ PKU, 
$^5$ CASIA, 
$^6$ SJTU, 
$^7$ EIT
\\
\small{$^{*}$: Equal Contribution, $\textsuperscript{\Letter}$ Corresponding authors}
\\
\vspace{-1pt}
\small{Project Page: \url{https://jiaweihe.com/dexvlg}}
}
\begin{document}

\twocolumn[{%
\maketitle
\centering
\vspace{-25px}
\begin{center}
    \includegraphics[width=0.96\textwidth]{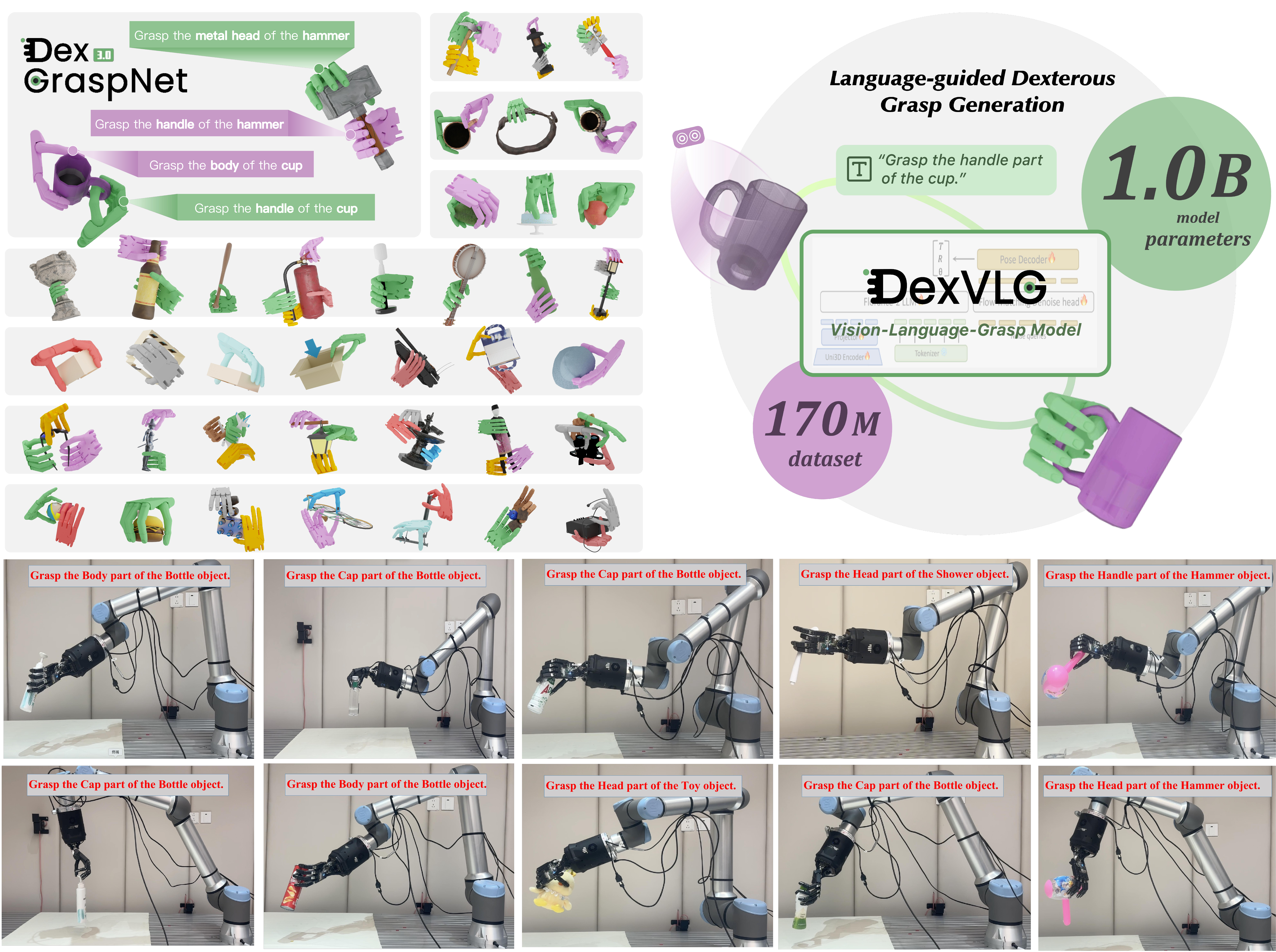}
    \vspace{-6px}
    \captionof{figure}{\textbf{Overview}. Our main contributions are twofold: First, we constructed a large-scale synthetic dexterous grasping dataset called DexGraspNet3.0, which contains grasp poses with captions describing the grasped part and style. Second, we trained a language-instructed grasp pose prediction model using the DexGraspNet3.0 dataset, called DexVLG. This model can generate language-aligned and generalizable grasping poses for different objects in real-world experiments.
    }
\label{fig:teaser}
\end{center}
\vspace{-2px}
}]

\begin{abstract}
As large models gain traction, vision-language-action (VLA) systems are enabling robots to tackle increasingly complex tasks.
However, limited by the difficulty of data collection, progress has mainly focused on controlling simple gripper end-effectors. There is little research on functional grasping with large models for human-like dexterous hands.
In this paper, we introduce \textbf{DexVLG}, a large \textbf{V}ision-\textbf{L}anguage-\textbf{G}rasp model for \textbf{Dexterous} grasp pose prediction aligned with language instructions using single-view RGBD input. To accomplish this, we generate a dataset of 170 million dexterous grasp poses mapped to semantic parts across 174,000 objects in simulation, paired with detailed part-level captions. This large-scale dataset, named \textbf{DexGraspNet 3.0}, is used to train a VLM and flow-matching-based pose head capable of producing instruction-aligned grasp poses for tabletop objects. To assess DexVLG's performance, we create benchmarks in physics-based simulations and conduct real-world experiments. Extensive testing demonstrates DexVLG’s strong zero-shot generalization capabilities—achieving over 76\% zero-shot execution success rate and state-of-the-art part-grasp accuracy in simulation—and successful part-aligned grasps on physical objects in real-world scenarios.
\end{abstract}
\section{Introduction}
To unleash the potential of intelligent abilities for a robot, recent advances in large vision-language-action (VLA) models~\cite{team2024octo,kim2024openvla,black2024pi_0} have shown a promising way. They have demonstrated strong generalizability on many complex robotic tasks across diverse scenarios in the real world. The key reason for their success is the large model capacity and training dataset: their model typically has billion-level parameters and is trained on billion-level robotic datasets. 

However, those large VLA models are currently limited to parallel grippers and cannot control a dexterous hand. The main reason is the lack of data for dexterous grasping. Some works retarget from human motion~\cite{yang2022oakink,wei2024grasp,liu2025dextrack} and teleoperate a real robot~\cite{wang2024dexcap,qin2023anyteleop} to collect data, but they all require significant human effort. Some works~\cite{wang2023dexgraspnet, zhang2024dexgraspnet, chen2024bodex} use analytic-based methods to quickly synthesize a large-scale dexterous grasp dataset, but they are semantic-unaware and thus cannot perform functional grasps like we humans do. For example, humans usually hold a hammer by the handle to use it, but may grip the metal part when handing it to someone else. Recent research on functional dexterous grasp~\cite{hang2024dexfuncgrasp, wu2024cross, huang2024fungrasp} can only use datasets with very limited scale, greatly limiting the model capacity and generalizability.

To address the data challenge, in this work, we propose a large-scale part-aware functional dexterous grasp dataset, named \textbf{DexGraspNet 3.0}. Our dataset contains 170M dexterous grasp poses on 174k objects from the Objaverse~\cite{deitke2023objaverse} dataset. Each grasp pose is validated in a physics-based simulation and paired with captions describing the grasped part name and grasp style. To build this, we follow the DexGraspNet series of works~\cite{wang2023dexgraspnet,zhang2024dexgraspnet} for efficient grasp synthesis, and introduce part-aware energies to make each grasp semantically distinguishable. We also leverage state-of-the-art object part understanding models like SAMesh~\cite{tang2024segment} and GPT-4o for part segmentation and captioning.

Powered by DexGraspNet 3.0, we developed \textbf{DexVLG}, a large vision-language-grasp model. DexVLG takes a language instruction and a single-view colored point cloud of tabletop objects as input and generates dexterous grasp poses based on the instruction.
DexVLG leverages multiple pre-trained foundation models to extract vision-language features and employs a flow-matching denoising paradigm to predict grasp poses. With billions of parameters, the model is fine-tuned end-to-end on our large-scale dataset.

To evaluate the performance of DexVLG, we perform experiments in both simulation and the real world. We first build a benchmark for part-aware dexterous grasping in Isaac Gym~\cite{makoviychuk2021isaac}, with novel metrics that evaluate the part-alignment of dexterous grasp poses. Several baselines are compared to show the superiority of our model. DexVLG outperforms baselines on all benchmarks and achieves over 76\% grasp success rate. We also demonstrate successful real-world executions predicted by DexVLG.

To summarize, our contributions are as follows:
\begin{itemize}
	\item We introduce DexGraspNet3.0, a large-scale dataset containing 170M part-aligned dexterous grasp poses on 174k objects, each annotated with semantic captions.
	\item We propose a VLM named DexVLG to generate language-instructed dexterous grasp poses in an end-to-end way.
	\item We curate benchmarks and conduct extensive experiments to evaluate DexVLG in simulation and the real world.
\end{itemize}
\begin{table*}[t]
    \small
    \centering
    \vspace{-20px}
    \begin{tabular}{l|l|ccccccc}
    \toprule
    \tabcolsep 1pt
    \textbf{Data Source}& \textbf{Dataset} & \textbf{Hand} & \textbf{Object} & \textbf{Grasp} & \textbf{Caption} & \textbf{Simulation Check} & \textbf{Semantics} & \textbf{Part}\\
    \hline
    \multirow{3}{*}{\textbf{Real-World}}
    & OakInk~\cite{yang2022oakink} & MANO & 100 & 50k & None & & $\surd$ & $\surd$ \\
    & DexGYSNet~\cite{wei2024grasp}& Shadow & 1800 & 50k & 50k & & $\surd$ & \\
    & SemGrasp~\cite{li2025semgrasp}& MANO & 1800 & 50k & 280k & $\surd$ & $\surd$ & \\
    \hline
    \multirow{5}{*}{\textbf{Simulation}} &
    Grasp'D~\cite{turpin2023fast} & Shadow,Allegro & 2408 & 1M & None & $\surd$ & &\\
    & DexGraspNet~\cite{wang2023dexgraspnet} & Shadow & 5355 & 1.32M & None & $\surd$ & &\\
    & DexGraspNet 2.0~\cite{zhang2024dexgraspnet} & Leap & 88 & 45.04M & None & $\surd$ & &\\
    & BoDex~\cite{chen2024bodex}& Shadow & 2397 & 7.17M & None & $\surd$ & &\\
    \cline{2-9}
    & \textbf{Ours} & \textbf{Shadow}   & \textbf{174k} & \textbf{170M} & \textbf{170M} & \textbf{$\surd$} & \textbf{$\surd$} & \textbf{$\surd$} \\
    \bottomrule
    \end{tabular}
    \vspace{-6px}
    \caption{\textbf{Comparison of DexGraspNet 3.0 with existing dexterous grasping datasets.} All of our grasp poses are validated in IsaacGym~\cite{makoviychuk2021isaac} and paired with part annotation and language captions.}
    \label{tab:datascale}
    \vspace{-6pt}
\end{table*}

\section{Related Work}
\subsection{3D Part Segmentation}
3D part segmentation splits a 3D object into distinct components. Early methods~\cite{loizou2023cross,zhao2021point} rely on human-annotated small datasets and struggle to generalize beyond the training distribution. The PartSLIP series~\cite{liu2023partslip,zhou2023partslip++} pioneers the application of 2D VLMs to 3D part segmentation. More recent works~\cite{yang2024sampart3d,ma2024find} pretrain 3D VLMs on the huge Objaverse~\cite{deitke2023objaverse} dataset and demonstrate much stronger generalizability. SAMesh~\cite{tang2024segment} is a zero-shot geometric part segmentation method on mesh, combining traditional shape geometric features and learning-based SAM. In this paper, we utilize the geometric SAMesh method to generate separated parts and VLMs~\cite{GPT4o24,gemini23} to assign semantics to parts.

\subsection{Dexterous Grasp Synthesis}
Many recent works on dexterous grasping are semantic-unaware. Some analytic-based synthesis works~\cite{liu2021synthesizing,turpin2022grasp, turpin2023fast,chen2024springgrasp,li2023frogger, chen2023task, chen2024bodex} study the efficient generation of grasps by optimizing a differentiable grasp quality metric, given the object mesh. These methods enable the generation of million-level dexterous grasp datasets like DexGraspNet 1.0~\cite{wang2023dexgraspnet} and 2.0~\cite{zhang2024dexgraspnet}.Others~\cite{jiang2021hand,chen2022learning,xu2023unidexgrasp, xu2024dexterousgrasptransformer,zhang2024dexgrasp} study to perform supervised learning on grasping datasets, using conditional generative models like CVAE, diffusion model and normalizing flows. Reinforcement learning~\cite{wan2023unidexgrasp++,zhang2024graspxl} is another hot topic for dexterous grasping, but up to now, most results are only presented in simulation.
A few works study semantic-aware dexterous grasping~\cite{agarwal2023dexterous,hang2024dexfuncgrasp,huang2024fungrasp,wu2024cross, li2025semgrasp,wei2024grasp}, but their studied object instances and categories are limited in scale. Our work, in contrast, introduces the semantic-aware energy to the advanced analytic-based method~\cite{zhang2024dexgraspnet, chen2023task, chen2024bodex}, synthesizing a large-scale functional dexterous grasp dataset.

\subsection{Vision-Language models for Robotic Action}
Using large vision-language models (VLM) to control robot action is an emerging research area. One way to achieve this is to decompose robot manipulation into a series of VQA tasks and execute 6D waypoints planned by a VLM~\cite{liu2024moka,wang2024vlm}. This is infeasible for dexterous grasping because a 6D end effector pose cannot fully characterize high-dimensional dexterous hand states. Another way is to learn generalizable action priors from large-scale robot trajectory datasets, which produces vision-language-action(VLA) models~\cite{team2024octo,kim2024openvla,brohan2023rt}. Existing dexterous grasp datasets are limited in scale and cannot support learning a VLA. The most relevant work to this paper is MultiGraspLLM~\cite{li2024multi}, which fine-tunes a VLM to predict dexterous hand pose tokens end-to-end. However, the generated grasp poses are not diverse enough.

\section{Notations and Task Specification}
We formulate the task of \textbf{language-instructed dexterous grasp generation} as follows: The input is a single-view colored point cloud $P$ of the object placed on a table, accompanied by a language instruction $T$ that specifies the semantic object part $S_i$ to be grasped and the grasping style. The details of language instructions are elaborated in \S\ref{section:caption}.\par
The output is a dexterous hand pose that correctly grasps the desired object part with the desired grasping style, as described in the input language instructions. A grasp is represented as $g = (T, R, \theta)$, where $T \in \mathbb{R}^3$ and $R \in \mathrm{SO}(3)$ define the wrist pose, and $\theta \in \mathbb{R}^d$ specifies the joint angles of the hand. We use the Shadow Hand, for which $d = 22$.

\section{DexGraspNet 3.0 Dataset}
\subsection{Dataset Statistics}
Table~\ref{tab:datascale} summarizes the key characteristics of the DexGraspNet 3.0 dataset. DexGraspNet 3.0 comprises 170 million dexterous grasps across 174k objects, making it the largest dexterous grasp dataset to date in terms of both grasp pose and object number. Each grasp is validated using the physics-based simulator IsaacGym~\cite{makoviychuk2021isaac} and paired with semantic captions and part-level annotations, resulting in 170M pose-caption pairs designed for training VLG models. The dataset visualization is shown in Fig.~\ref{fig:visdataset}.

\subsection{Object Preparation and Part Segmentation}
Objects are sourced from the Objaverse~\cite{deitke2023objaverse} dataset and filtered using GPT-4o~\cite{achiam2023gpt}, following~\cite{sofar25}. The assets are then processed with ManifoldPlus~\cite{huang2020manifoldplus} and CoACD~\cite{wei2022approximate} to generate collision meshes, yielding 229K valid objects. For each valid object, GPT-4o~\cite{achiam2023gpt} estimates a reasonable size, and normalization is performed accordingly (see Appendix Sec. B for details).

\begin{figure*}[h]
    \centering
    \vspace{-15pt}
    \includegraphics[width=\linewidth]{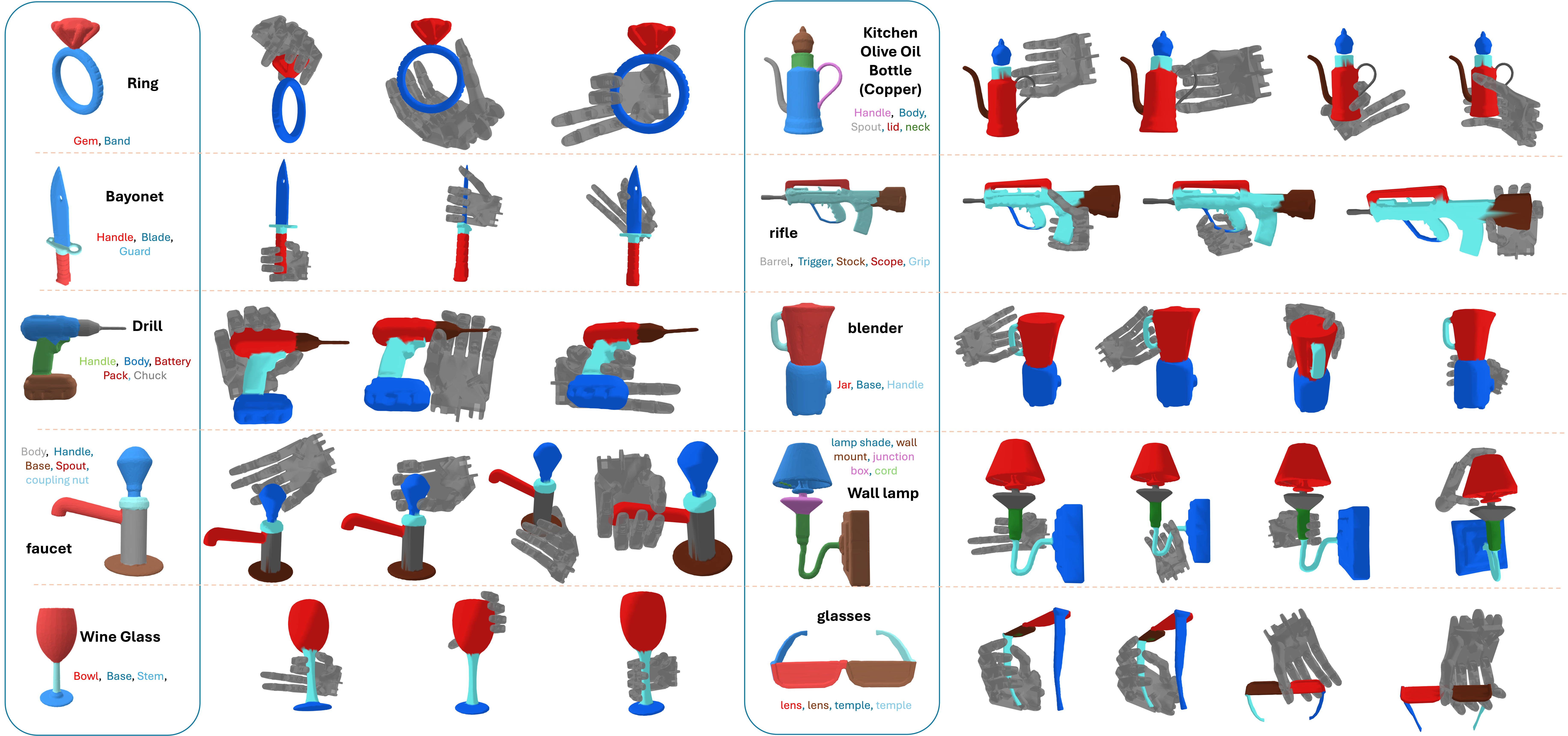}
    \vspace{-15px}
    \caption{\textbf{Visualization of part-aware dexterous grasp poses in DexGraspNet3.0}. The left columns visualize sample objects together with part segmentation generated by SAMesh~\cite{tang2024segment} and captioned by GPT-4o~\cite{GPT4o24}. On the right are part-aligned grasp poses generated by our optimization pipeline. Each grasp makes contact with a single object part and naturally aligns with the way humans grasp objects.}
    \label{fig:visdataset}
    \vspace{-3px}
\end{figure*}

We use SAMesh~\cite{tang2024segment} to perform zero-shot geometry-based part segmentation on colorless collision meshes. Fig.~\ref{fig:visdataset} presents visualizations of the segmentation results, providing sufficient functional priors. The part-segmented objects are rendered from multiple views, and part names are automatically labeled using set-of-mark prompting~\cite{yang2023set} with GPT-4o.

\subsection{Part-aware Dexterous Grasp Generation}
\begin{figure*}[h]
    \centering
    \includegraphics[width=\linewidth]{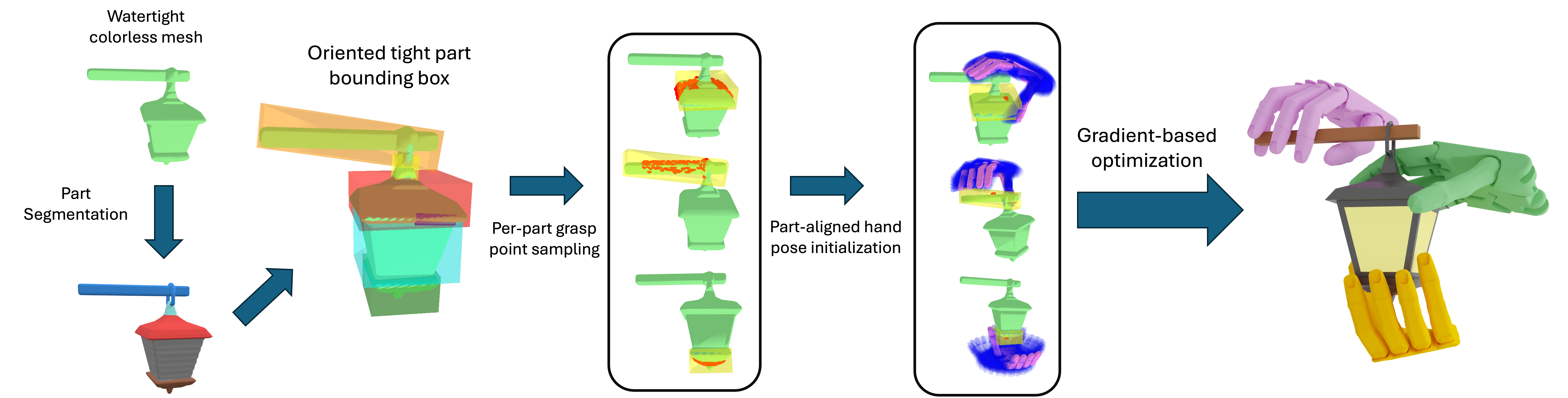}
    \vspace{-16pt}
    \caption{\textbf{Grasp pose generation pipeline}. Given a watertight colorless mesh of an object, we perform part segmentation with SAMesh~\cite{tang2024segment} and fit oriented tight bounding boxes for each part. On each part, we sample grasp points on certain areas and align hand initialization poses with the part by leveraging geometric cues in bounding box parameters and grasp points. We further jitter each initial hand pose and run batched gradient-based optimization, which generates final dexterous grasp poses on each object part.}
    \label{fig:datapipe}
    \vspace{-3px}
\end{figure*}

Our grasp synthesis pipeline is shown in~\cref{fig:datapipe}. It is built upon the advanced analytic-based method~\cite{zhang2024dexgraspnet, chen2024bodex}, which uses cuRobo~\cite{sundaralingam2023curobo} to support massive parallelization on GPUs. To adjust the previous semantic-unaware pipeline for us, we propose a part-aware hand pose initialization strategy and several energy functions, as introduced below. 

\subsubsection{Part-Aware Hand Pose Initialization}
The initial hand pose is regarded as greatly affecting the result of gradient-based optimization, as observed in DexGraspNet~\cite{wang2023dexgraspnet}. Although that work proposes an initialization method, it is not suitable to our scenario, which requires grasping a specific part. 
As shown in Fig.~\ref{fig:datapipe}, we first generate the oriented bounding boxes (OBB) of object parts and sample grasp points from certain areas on the part surface. Then, we set the palm pose and initial joint angles using rules that rely on the geometric cues indicated by the OBB. The wrist pose is further randomly jittered to obtain a diverse distribution. More details are in the Appendix. 

\begin{figure*}[t!]
      \begin{center}
      \vspace{-10pt}
       \includegraphics[width=0.88\linewidth]{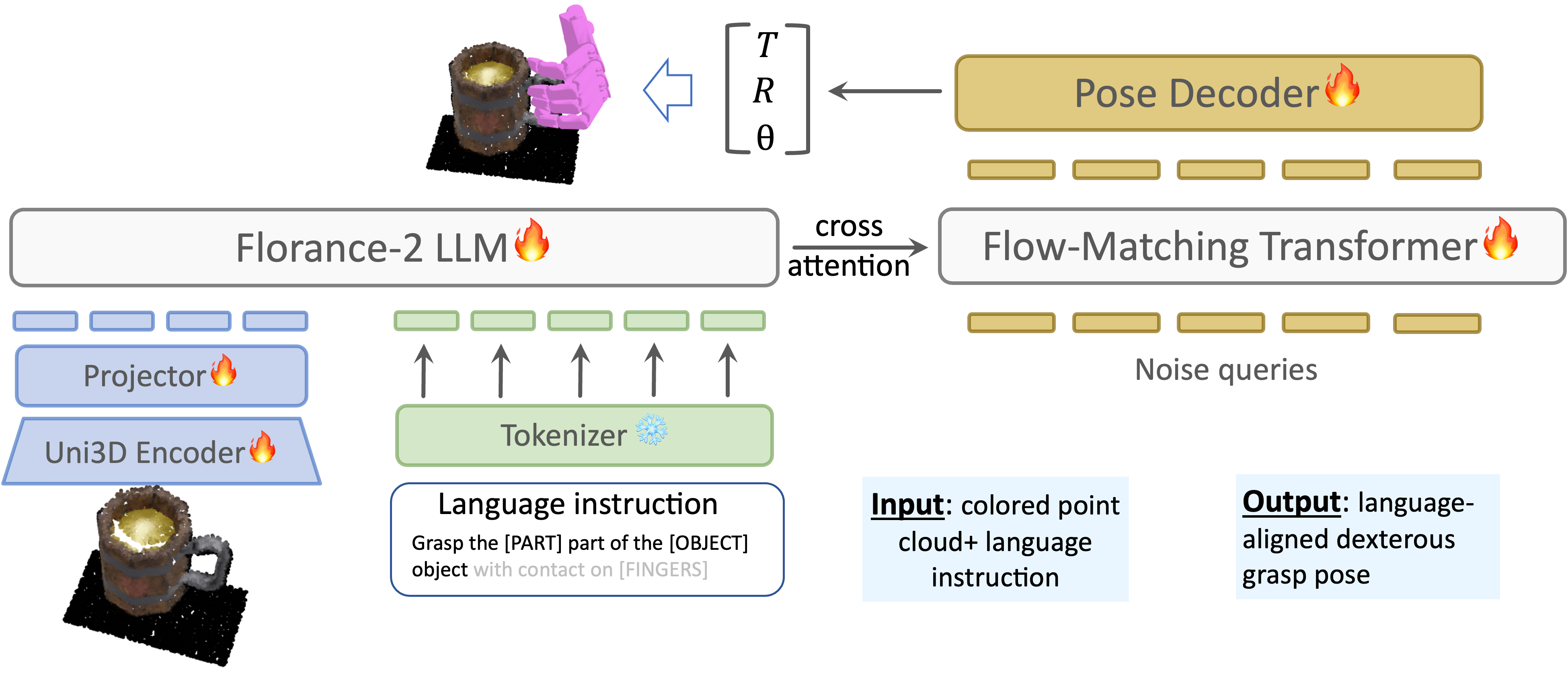}
       \end{center}
        \vspace{-16pt}
         \caption{\textbf{Pipeline of our DexVLG model.} It first uses Uni3D~\cite{zhou2024unid} to encode the colored point cloud. Then, Florence-2~\cite{xiao2024florence} is leveraged to reason the projected visual features and the language tokens. Next, a flow-matching denoise head will conditioned on the LLM output embeddings and finally, a pose decoder will output a desired grasp. Note that all of our modules, except the language tokenizer, are fine-tuned end-to-end using our DexGraspNet 3.0 dataset.}
          \vspace{-5pt}
         \label{fig:dexvlg}
\end{figure*} 

\subsubsection{Objectives for Gradient-based Optimization}
In this section, we formulate the physics-based energy functions used for gradient-based optimization.

\textbf{LP-based differentiable force closure energy} $E_{FC}$.
Many recent works~\cite{liu2021synthesizing, li2023frogger, zhang2024dexgraspnet, chen2024bodex} propose different kinds of differentiable force closure metrics to evaluate the grasp quality. We adopt the variant proposed in DexGraspNet 2.0~\cite{zhang2024dexgraspnet} to balance the speed and performance. On the one hand, our LP-based energy uses linear programming (LP) to adjust the contact forces, relaxing the assumption of equal contact force in the original DFC metric~\cite{liu2021synthesizing, wang2023dexgraspnet}. On the other hand, our energy assumes no friction, which avoids the heavy computation of quadratic programming as in BODex~\cite{chen2024bodex}.
More details are in the Appendix. 

\textbf{Part-contact energy }
To encourage a better contact convergence between the hand and the desired part, 
We follow the collision detection algorithm in the IPC simulator~\cite{li2020incremental} and define a truncated barrier function $E_{bar}$ that repulses fingertips from object surfaces outside the target part:
\begin{equation}
    E_{bar} = \sum\limits_{n=1}^5\sum\limits_{j\neq i} b(d(x_n, p_j), d_{thr})
\end{equation}
\begin{equation}
    b(d,d_{thr}) = \left\{\begin{aligned}
        &-(d-d_{thr})^2ln\left(\frac{d}{d_{thr}}\right), & 0<d<d_{thr} \\
        &0, &d\geq d_{thr}
    \end{aligned}\right.
\end{equation}

Where $\{x_n\}_{n=1}^5$ are the fingertips, $p_j$ are point clouds sampled from object surface outside the target part $s_i$. $E_{bar}$ goes to infinity when any of the fingers make contact with an object outside part $s_i$, hence strictly enforces part alignment when the stepsize is small enough.

\textbf{Distance energy} We minimize the distance between fingertips and the object to ensure contact and encourage the palm contact points to keep a distance $d_0=1cm$ from the object.
\begin{equation}
    E_{dis} = \sum\limits_{i=1}^n d(x_n, O) + \omega_{palm} |d(x_{palm}, O) - d_0|
\end{equation}

Besides, we implement several regularization energies, aggregated as $E_{reg}$, to prevent hand-object collision, hand self-collision and encourage the contact points to align with the front side of fingers. The complete energy function is
\begin{equation}
    E = \omega_{FC}E_{FC} + \omega_{bar}E_{bar} + \omega_{dis}E_{dis} + \omega_{reg}E_{reg}
\end{equation}

\subsection{Grasp Validation and Captioning}
\label{section:caption}
To obtain a high-quality dataset free of undesirable poses after optimization is complete, we validate all the final poses with a physics-based simulator IsaacGym~\cite{makoviychuk2021isaac}. In particular, we check against a set of 4 criteria and only consider grasp poses that pass all of them as valid: 1. penetration between the hand and object is less than 3mm; 2. The self-penetration distance of the hand is less than 3mm; 3. It counteracts gravity from all six axis-aligned directions in simulation; 4. the \textbf{part-alignment condition}, which indicates that if a hand link is in contact with the object (i.e. the distance to the object is less than 0.2 cm), this hand link should be closer to the desired part than any other parts of the object.

We caption each grasp pose with the template \textbf{``Grasp the \{part\} of the \{object\} object, with contacts on \{fingers\}"}, where \{part\} and \{object\} are the part name and object name inferred by GPT-4o. The part-alignment condition ensures that each grasp pose has a meaningful part name in correspondence. 
\{fingers\} lists the names of all fingers in contact with the object part,  which is checked in simulation. Each caption contains rich semantic and contact information for the model to learn.

\subsection{Table-top Scene Generation and Rendering}
The above grasping poses are generated for floating objects, but the object is often placed on a table in the real world. Therefore, we also need to generate diverse poses for objects being stably placed on a table. Following Open6DOR~\cite{ding2024open6dor}, we uniformly sample N=1000 initial rotations from SO(3) and drop the object from a height of 10cm onto the ground. We simulate for 5 seconds and all stabilized poses are collected and deduplicated. Then we transform the grasp poses using the generated object poses and filter out those grasps that have collisions with the table. 
Each scene is rendered from eight views using a RealSense D415 RGBD camera in Blender. The camera poses are visualized in the Appendix.

\section{DexVLG Model}
We propose a large vision-language-grasp model, called DexVLG, to tackle the task of language-instructed dexterous grasping. As illustrated in Fig.~\ref{fig:dexvlg}, DexVLG tasks a single-view point cloud observation and a language instruction as input, and outputs a grasp pose that satisfies the language instruction.

\subsection{Point Cloud (PC) Encoder}
The PC encoder takes colored point clouds from a single view as input. There are lots of foundational PC encoders~\cite{act23,recon23,vpp23,qi2025shapellm} from pretraining. We adopt the pretrained Uni3D~\cite{zhou2024unid} backbone, which has a ViT~\cite{dosovitskiy2021an}-based architecture scaled from small (23M) to large (307M). Uni3D~\cite{zhou2024unid} is pretrained to align point cloud features with CLIP~\cite{radford2021learning} features via contrastive learning~\cite{zhou2024unid}, therefore has the ability to extract semantic information from raw point clouds. The point cloud is downsampled into a fixed number $n_p=10000$ with furthest point sampling before being given to the encoder. The encoded 3D features are then fed into an MLP projector to align the PC features with pretrained large language models.

\subsection{Language Foundation Model}
We adopt the LLM base model and language tokenizer from Florence-2~\cite{xiao2024florence}, which varies in model parameter size from Base (232M) to Large (771M).
We concatenate PC features with language embeddings to create the input for the large language model.
Following Transfusion~\cite{zhou2025transfusion} and $\pi_0$~\cite{black2024pi_0}, the LLM will share the cross-attention with the flow-matching-based pose prediction head.

\subsection{Flow Matching-based Grasp Generation}
We use the flow matching-based pose denoising module to generate the dexterous grasp pose, which is learned by minimizing the mean square objective: 
\begin{equation}
\begin{aligned}
\label{eq:rect_flow_obj}
    \min_{v} \mathbb{E}_{(t,X_0, X_1)\sim \gamma} \mid \mid \frac{\mathrm{d}}{\mathrm{d} t} X_t  - v(X_t, t) \mid \mid^2 
\end{aligned}
\end{equation} 

where $X_t = \phi(X_0,X_1,t)$ is any time-differentiable interpolation between ground truth grasp pose $X_1$ and a sample $X_0$ from noise distribution,
with $\frac{\mathrm{d}}{\mathrm{d} t} X_t  = \partial_t \phi(X_0,X_1,t)$. The denoising process is conditioned on the LLM's hidden states, with the denoise module sharing transformer architecture with the LLM. An MLP serves as the pose decoder, generating the grasp pose by determining the hand base's translation and rotation along with the joint angles of each finger.

\section{Experiments}
We evaluate DexVLG against baseline models to answer the following questions:

\textbf{Does DexVLG zero-shot generate high-quality dexterous grasp poses on a variety of objects and semantic parts?} 
\textbf{How well do DexVLG-predicted grasp poses align with language instructions?} We define metrics that evaluate the quality of grasp poses and how well the poses align with language instruction, and test DexVLG in diverse benchmarks in simulation.

\textbf{Is it necessary to use large vision-language models to address the task of instruction-aligned dexterous grasp generation?} We compare DexVLG against baselines implemented with small model, and analyze the benefits of leveraging large vision-language model.

\subsection{Training Details}
The entire DexVLG model undergoes single-stage full-parameter fine-tuning, with only the language tokenizer frozen and all other modules updated from end-to-end fine-tuning. The learning rate is set to $6\times10^{-5}$, with a warm-up strategy applied for the first 3 epochs. Weight decay is $1\times10^{-4}$. The Adam optimizer is used for training over 230 epochs on 64 NVIDIA RTX 4090 GPUs. Here, we define an epoch as sampling a grasp pose for a single part.

\subsection{Evaluation Metrics}
The following metrics are used to evaluate the \textbf{quality} and \textbf{instruction alignment accuracy} of grasp poses generated by different models in simulation.

\textbf{Simulation Success Rate (Suc)} represents the percentage of successful grasp executions in simulation, defined as lifting the object up 3cm from the table and holding for 1s. Our implementation follows the protocol in~\cite{zhang2024dexgraspnet}.

\textbf{Part Touch Accuracy (PTA)} represents the percentage of grasp poses that touch the target semantic part, evaluated by checking whether at least one finger in the predicted pose is less than 1cm away from the part and is closer to the desired part than any other part.

\textbf{Part Grasp Accuracy (PGA)} represents the percentage of grasp poses that form a grasping pose at the target part, evaluated by checking whether at least three fingers in the predicted pose are less than 1cm away from the part and are closer to the desired part than any other part.

\subsection{Simulation Benchmark Result}
We compare DexVLG against DexGraspNet2.0~\cite{wang2023dexgraspnet}, a diffusion-based small model for dexterous grasp generation in tabletop scenes. We retrain DGN2.0 on DexGraspNet3.0 dataset, dubbed \textbf{DGN2.0$^{*}$}. To further equip DGN2.0 with language understanding ability, we concatenate the CLIP~\cite{radford2021learning} embedding of text instructions with point cloud features, and supervise with part-specific ground truth labels. We dub this augmented baseline \textbf{DGN2.0$^{*}$+CLIP}.

\begin{figure*}[h]
    \centering
    \includegraphics[width=\linewidth]{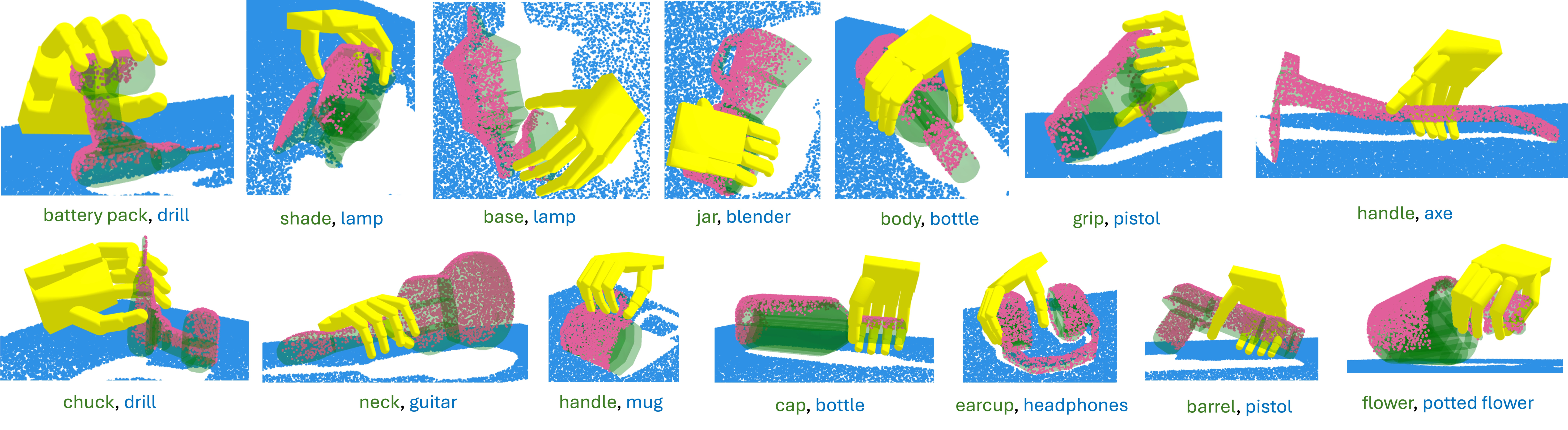}
    \vspace{-20px}
    \caption{\textbf{Visualization of grasp poses predicted by DexVLG in simulation.} The object mesh is drawn only for visualization. The model input is a single-view point cloud, the color present in this figure is painted only for visualization. Each grasp is instructed with "Grasp the \textcolor{green}{[PART]} part of \textcolor{blue}{[OBJECT]} object".}
    \label{fig:visdata}
    \vspace{-3px}
    
\end{figure*}

\begin{table}[h]
    \centering
    \begin{tabular}{l|l|c|c|c}
    \toprule
    \multirow{2}{*}{Benchmark} & \multirow{2}{*}{Metric}
    &\multirow{2}{*}{DGN2.0$^*$}
    &DGN2.0$^*$
    &\multirow{2}{*}{\textbf{Ours}}\\
    & & & +CLIP & \\
    \midrule
    \multirow{3}{*}{LVIS-Seen}&\textbf{Suc}$\uparrow$&70.8&67.8&\textbf{87.7}\\
    &\textbf{PTA}$\uparrow$&54.0&55.2&\textbf{70.7}\\
    &\textbf{PGA}$\uparrow$&38.5&40.5&\textbf{62.1}\\
    \midrule
    \multirow{3}{*}{Unseen}&\textbf{Suc}$\uparrow$&57.7&56.0&\textbf{79.1}\\
    &\textbf{PTA}$\uparrow$&63.9&64.7&\textbf{68.2}\\
    &\textbf{PGA}$\uparrow$&25.3&26.1&\textbf{36.3}\\
    \midrule
    \multirow{3}{*}{SamPart3D}&\textbf{Suc}$\uparrow$&56.8&55.1&\textbf{76.3}\\
    &\textbf{PTA}$\uparrow$&49.2&50.8&\textbf{66.0}\\
    &\textbf{PGA}$\uparrow$&38.8&39.4&\textbf{52.0}\\
    
    \bottomrule
    \end{tabular}
    \vspace{-5px}
    \caption{\textbf{Result on simulation benchmarks}.}
    \label{tab:bench-lvis}
    \vspace{-5px}
\end{table}

\begin{table}[h]
    \centering
    \begin{tabular}{l|ccc}
    \toprule
    Method & \textbf{Suc}$\uparrow$ & \textbf{Part Grasp}$\uparrow$ & \textbf{CMA}$\uparrow$\\

    \midrule
    DGN2.$0^{*}$+CLIP&68.2&35.2&21.8\\
    + contact label&72.3&33.8&23.0\\
    \midrule
    Ours   &73.0&33.9&27.0 \\
    \textbf{+ contact label}   &\textbf{76.1}&\textbf{48.1}&\textbf{29.8} \\
    \bottomrule
    \end{tabular}
    \vspace{-5px}
    \caption{\textbf{Result of contact mode learning on LVIS-Seen benchmark in simulation}. The \textbf{contact mode accuracy (CMA)} metric refers to contact mode accuracy, which reflects the rate of grasp poses that match the instructed contact mode, allowing differences with at most one finger contact.}
    \label{tab:bench-contact}
    \vspace{-3px}
\end{table}

To evaluate the models, we curate three benchmarks in simulation. In each benchmark, we manually select well-segmented objects and filter their poses on table such that the target parts for grasping are not occluded by the table. 
\begin{itemize}
    \item The \textbf{LVIS-Seen} benchmark consists of 40 seen objects in the Objaverse-LVIS split.
    \item The \textbf{Unseen} benchmark consists of 84 Objaverse objects unseen in the training process.
    \item The \textbf{SamPart3D} benchmark consists of 56 Objaverse objects segmented and semantically annotated by SamPart3D~\cite{yang2024sampart3d} using methods different from our work.
\end{itemize}

\subsubsection{Part-conditioned Grasp Generation}
In this experiment, we train and infer with input language instruction ``\textbf{Grasp the \{part\} part of the \{object\} object}". The quantitative results are shown in Tab.~\ref{tab:bench-lvis}. Our DexVLG outperforms DGN2.0+CLIP in terms of both simulation success rate and part accuracy in all benchmarks, demonstrating superior performance. DexVLG robustly generalizes to grasping unseen objects and retains 79\% success rate and 36\% part grasp accuracy. It also generalizes with respect to part segmentation methods and retains a 76\% success rate and a 52\% part grasp accuracy in the SamPart3D benchmark. The qualitative results are shown in Fig.~\ref{fig:visdata}.

It is worth noting that even though the Florence-2 LLM backbone~\cite{xiao2024florence} is not pretrained on object part understanding tasks and haven't learned about the novel part names of objects in \textbf{Unseen} benchmark through finetuning, the DexVLG model still learns to follow instruction of grasping these parts and achieve substantially higher part grasp accuracy than baselines. This result demonstrates the strong generalizability of VLMs in learning language-aligned dexterous grasping tasks.

Interestingly, we find that augmenting the DGN2.0 baseline with CLIP features enhances its language alignment ability at the cost of harming the quality of generated poses, reflected in a drop in simulation success rate. This trade-off behavior indicates \textbf{small models are limited in capacity, which hinders them from learning complex tasks such as generating instruction-aligned dexterous grasps.} On the other hand, Tab.~\ref{tab:ablation_caption} shows training with language condition enhances the performance of DexVLG, which indicates \textbf{the large capacity of VLM is necessary for learning instruction-aligned dexterous grasp generation.}

\subsubsection{Contact Mode-conditioned Grasp Generation}
In this experiment, we benchmark different models on a more complex instruction following task: inference with language instruction ``\textbf{Grasp the \{part\} part of the \{object\} object with contact on \{fingers\}}", where \{fingers\} are the names of fingers that we want to make contact with the object. We compare models trained with \textbf{ (dubbed ``+contact label")} and without contact mode labels on the \textbf{LVIS-Seen} benchmark. The quantitative results are shown in Tab.~\ref{tab:bench-contact}. Models trained without a contact mode label are confused by this extra instruction and degenerate in performance. On the other hand, when trained with contact mode annotation, DexVLG effectively learn to follow the extra instruction and improves performance. This experiment demonstrates \textbf{large VLMs are capable of learning more complex instruction-following grasp generation tasks, which small model struggles to learn}.

\subsection{Real-World Experiments}
Our real-world experiments utilize a ShadowHand mounted on UR10e robotic arm. An Intel RealSense D415 camera is mounted at the wrist and captures static single-view colored point cloud from a 45-degree lookdown perspective. During the experiment, we prompt DexVLG with language instruction specifying desired parts, such as ``grasp the neck of the bottle". The generated poses are filtered against safety constraints and executed with motion planning. We report \textbf{Grasp Success Rate} as the rate of successfully lifting up, and report \textbf{Part Accuracy} by human examination. DexVLG achieves 80\% success rate and 75\% part accuracy with these simple objects. Limited by hardware workspace and safety concerns, we do not report real-world results with more complex objects.

\begin{figure}[t!]
  \begin{center}
  \includegraphics[width=1.0\linewidth]{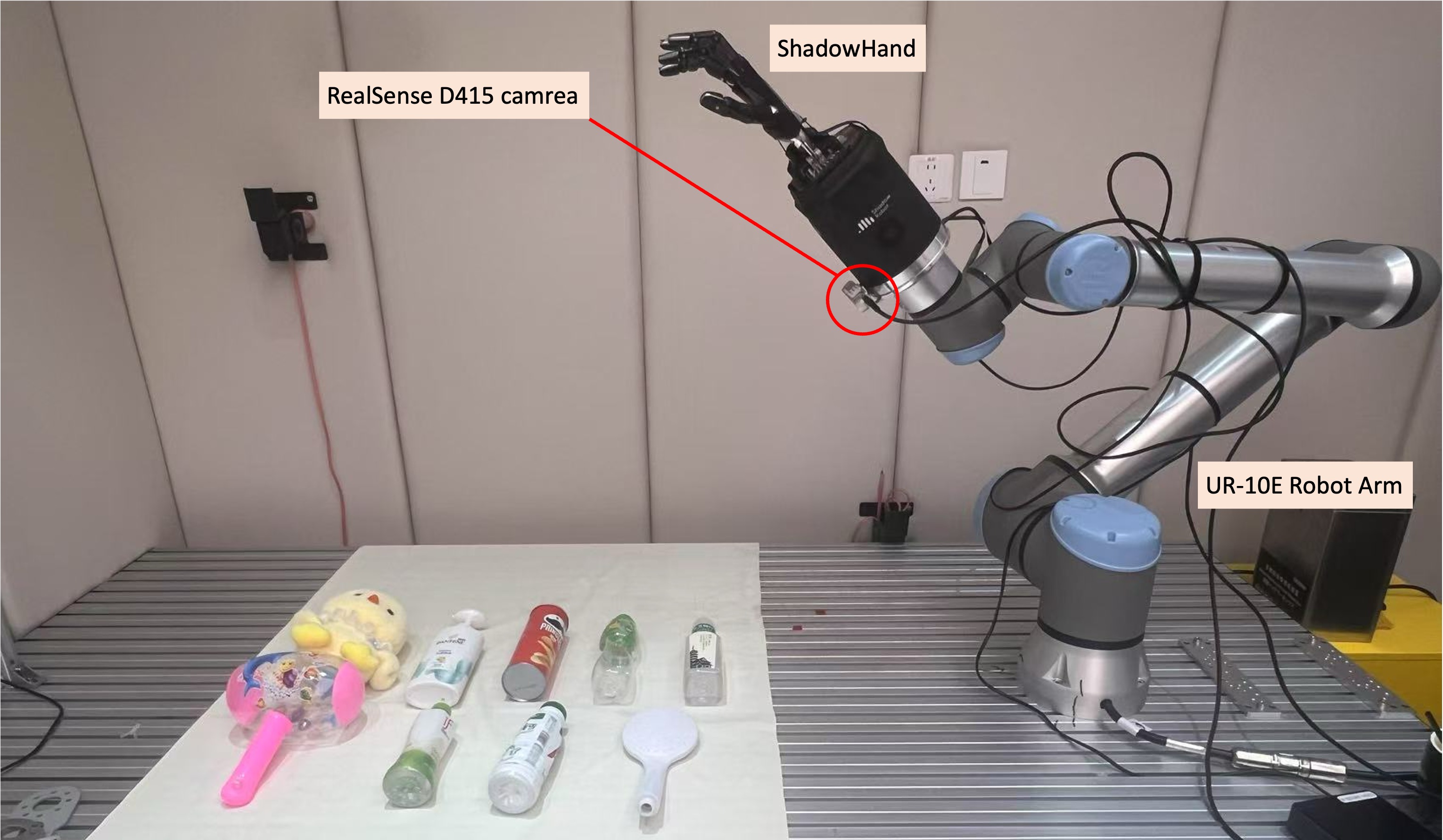}
  \vspace{-16px}
  \caption{\textbf{real-world experiment setting}.} \label{fig:real_setting}
  \end{center}
  \vspace{-16px}
\end{figure}

\subsection{Ablation Study}
We conduct ablation studies on model architecture, data scale, and input information. As shown in Table~\ref{tab:abl_data}, training data scale plays a crucial role in performance, with a 10\% to 100\% increase significantly improving success rate (Suc) and part grasp accuracy (PGA) across all datasets. The most substantial gains occur in unseen objects and 3D part recognition, emphasizing the importance of larger datasets for better generalization.
Our model size study (Table~\ref{tab:abl_model}) shows that increasing parameters from 255M to 1B provides only marginal improvements in success rate, with inconsistent effects on part accuracy. The 320M model performs best for part accuracy on LVIS-Seen and Unseen, while the 1B model excels in overall success rate, especially for SamPart3D. This suggests that larger models do not always enhance part-level understanding, and a mid-sized model (320M) balances efficiency and performance.
Our input information study (Table~\ref{tab:abl_color}) reveals that adding color to point clouds (PC w/ color) significantly improves both success rate and part grasp accuracy. The biggest gains appear in LVIS-Seen (Suc: +27.7, Part G: +23.5) and SamPart3D (Suc: +19.8, Part G: +13.1). These findings highlight the critical role of color in enhancing both object-level and part-level understanding in 3D.
\begin{table}[h]
    \centering
    \resizebox{\columnwidth}{!}{\begin{tabular}{l|cc|cc|cc}
    \toprule
    \multirow{2}{*}{Data}  &
    \multicolumn{2}{c|}{LVIS-Seen} & \multicolumn{2}{c|}{Unseen} & \multicolumn{2}{c}{SamPart3D}\\
    &  \textbf{Suc}$\uparrow$ & \textbf{PGA}$\uparrow$ & \textbf{Suc}$\uparrow$ & \textbf{PGA}$\uparrow$& \textbf{Suc}$\uparrow$ & \textbf{PGA}$\uparrow$\\
    \midrule
   with instruction &87.7 &62.1 & 79.1 & 36.3 & 76.3 & 52.0\\
   no instruction &84.4 &-- &64.5&--&70.2&--\\
    \bottomrule
    \end{tabular}
    }
    \vspace{-6px}
    \caption{\textbf{Ablation study for input language instruction}.}
    \label{tab:ablation_caption}
    \vspace{-7px}
\end{table}
\begin{table}[h]
    \centering
    \resizebox{\columnwidth}{!}{\begin{tabular}{l|cc|cc|cc}
    \toprule
    \multirow{2}{*}{Data}  &
    \multicolumn{2}{c|}{LVIS-Seen} & \multicolumn{2}{c|}{Unseen} & \multicolumn{2}{c}{SamPart3D}\\
    &  \textbf{Suc}$\uparrow$ & \textbf{PGA}$\uparrow$ & \textbf{Suc}$\uparrow$ & \textbf{PGA}$\uparrow$& \textbf{Suc}$\uparrow$ & \textbf{PGA}$\uparrow$\\
    \midrule
   10\% &49.7 &12.5 &32.3 &7.9 &28.4 &11.7\\
   20\% &75.3 &39.1  &54.0 &18.3 &53.4 &27.0\\
   100\%&87.7 &62.1 &79.1 &36.6 &76.3 &52.0\\
    \bottomrule
    \end{tabular}
    }
    \vspace{-6px}
    \caption{\textbf{Ablation study on training data scaling}. We explore the data efficiency from 10\% (about 17k objects) to 100\%.}
    \label{tab:abl_data}
    \vspace{-7px}
\end{table}
\begin{table}[h]
    \centering
    \resizebox{\columnwidth}{!}{\begin{tabular}{l|cc|cc|cc}
    \toprule
         \multirow{2}{*}{Param} &
    \multicolumn{2}{c|}{LVIS-Seen} & \multicolumn{2}{c|}{Unseen} & \multicolumn{2}{c}{SamPart3D}\\
     &\textbf{Suc}$\uparrow$ & \textbf{PGA}$\uparrow$ & \textbf{Suc}$\uparrow$ & \textbf{PGA}$\uparrow$& \textbf{Suc}$\uparrow$ & \textbf{PGA}$\uparrow$\\
    \midrule
     255M&74.8 &35.6 &55.2 &16.4 &47.6 &26.5\\
     320M&72.3 &41.6 &53.7 &20.5 &50.0 &22.8\\
     1B &75.3 &39.1 &54.0 &18.3 &53.4 &27.0\\
    \bottomrule
    \end{tabular}
    }
    \vspace{-6px}
    \caption{\textbf{Ablation study for model size}. The 225M model uses Florence-2-base and Uni3D-small, 320M for Florence-2-base and Uni3D-base, and 1B for Florence-2-large and Uni3D-large.}
    \label{tab:abl_model}
    \vspace{-7px}
\end{table}
\begin{table}[h]
    \centering
    \resizebox{\columnwidth}{!}{\begin{tabular}{l|cc|cc|cc}
    \toprule
       \multirow{2}{*}{Input}  &
    \multicolumn{2}{c|}{LVIS-Seen} & \multicolumn{2}{c|}{Unseen} & \multicolumn{2}{c}{SamPart3D}\\
    &  \textbf{Suc}$\uparrow$ & \textbf{PGA}$\uparrow$ & \textbf{Suc}$\uparrow$ & \textbf{PGA}$\uparrow$& \textbf{Suc}$\uparrow$ & \textbf{PGA}$\uparrow$\\
    \midrule
   PC w/o color &47.6 &15.6 &35.0 &10.4 &33.6 &13.9\\
   PC w/ color&75.3 &39.1 &54.0 &18.3 &53.4 &27.0\\
    \bottomrule
    \end{tabular}
    }
    \vspace{-6px}
    \caption{\textbf{Ablation study on input information}. We compare the impact of using colored point clouds.}
    \label{tab:abl_color}
    \vspace{-7px}
\end{table}

\section{Limitations and Conclusions} 
In this paper, we present DexVLG, an end-to-end language-aligned dexterous grasp generation model that leverages the capacity of large VLMs, trained with our synthesized large-scale DexGraspNet3.0 dataset. DexVLG achieves state-of-the-art performance in both grasp success and part accuracy in simulation, and achieves 80\% success rate in grasping simple objects in the real world. Nonetheless, our work has several limitations. As training poses in the DexGraspNet3.0 dataset is synthesized with floating hands without considering the hand-arm workspace, many poses sampled by DexVLG are unsafe to execute in real-world. As another limitation, our method does not support effective ranking of generated grasp poses. Ranking large batches of samples by likelihood score~\cite{grathwohl2018ffjordfreeformcontinuousdynamics}, as done in~\cite{zhang2024dexgraspnet}, is infeasible for VLM-based models as retaining gradients with respect to VLM parameters costs huge GPU memory. We leave developing effective sample ranking methods as future work.



{
    \small
    \bibliographystyle{ieeenat_fullname}
    \bibliography{references}
}
\clearpage
\appendix
\section{Implementation Details of Energy-based Optimization}
\label{sec:DFC}

\subsection{Preliminary: the Force-Closure property of dexterous grasp poses}
The concept of \textbf{force closure} is first proposed in~\cite{ferrari1992planning} as a physics-based property desirable for multi-contact grasp poses on objects. Under the assumption of the Coulomb friction model~\cite{ferrari1992planning}, force closure is a binary metric indicating whether a grasp pose can resist external wrenches applied to the object and hold the object in a static status:

\hypbox{Definition of \textbf{Force Closure}}{
If a set of contact points \{$C_m$,$n_m$\} in a Coulomb friction model is force closure, where $C_m$ are the positions of contact and $n_m$ are the contact normals pointing to the inside of object, then for any external force $F$ applied on the object, there exists contact forces \{$f_m$\} that lies within the friction cones correspond to \{$C_m$,$n_m$\} and cancel out the external force.
}

We refer the readers to the paper~\cite{ferrari1992planning} for a rigorous formulation and more technical details. The concept of force closure offers a handy metric to check whether a dexterous grasp pose is stable. However, it is not clear from the definition \textbf{how we can optimize a dexterous grasp pose towards forming a force closure grasp}. Many existing works explore approximation methods and formulate force closure into objective functions of optimization problems~\cite{5354501, mahler2017dexnet20deeplearning,dai2018synthesis}. In this paper, we follow~\cite{liu2021synthesizing} and base our method on the differentiable force closure (DFC).

\subsection{Preliminary: Differentiable Force Closure (DFC)}
\citep{liu2021synthesizing} first proposes a differentiable force closure estimator (DFC) as an energy term applicable to gradient-based optimization for dexterous grasp synthesis. Given a collection of $n$ contact force vectors, which is approximated by $n$ contact normal vectors $c\in\mathbb{R}^{3n}$ at $n$ positions $x\in\mathbb{R}^{3n}$, DFC encourages them to form a force closure over the grasped shape. Formally, DFC is expressed as
\begin{equation}
    E_{fc} = \Vert Gc\Vert_2
\end{equation}

where

\begin{equation}
    c = \begin{bmatrix}
        c_1, ..., c_n
    \end{bmatrix} \in \mathbb{R}^{3n}
\end{equation}

\begin{equation}
    G= \begin{bmatrix}
        I_{3\times3}& ... & I_{3\times3}\\
        [x_1]_{\times}& ... & [x_n]_{\times}
    \end{bmatrix} \in \mathbb{R}^{6\times3n}
\end{equation}

\begin{equation}
    [x_i]_{\times}= \begin{bmatrix}
        0 & -x_i^{(z)} & x_i^{(y)} \\
        x_i^{(z} & 0 & -x_i^{(x)} \\
        -x_i^{(y)} & x_i^{(x)} & 0
    \end{bmatrix} \in \mathbb{R}^{3\times3}
\end{equation}
$$x_i = (x_i^{(x)},x_i^{(y)},x_i^{(z)}) \in \mathbb{R}^3$$
Intuitively, the matrix $G$, named the \textit{grasp matrix}, transforms a set of contact forces in unit magnitude $c$ into net external force and torque applied to the object. As analyzed in~\cite{dai2018synthesis,liu2021synthesizing}, a small value of $\Vert Gc\Vert_2$ indicates the set of contact forces effectively cancel out each other without considering the effect of friction, which suggests the contact force vectors span the entire wrench space and therefore can potentially counteract any external wrench applied to the object. As a result, optimizing with $E_{fc}$ encourages synthesizing force-closure grasp poses.

By using surface normal vectors to approximate contact forces, DFC assumes equal magnitude contact forces at all contact points. For more details, please refer to ~\cite{liu2021synthesizing}.

\begin{figure}[h]
    \centering
    \vspace{-10pt}
    \includegraphics[width=1.0\linewidth]{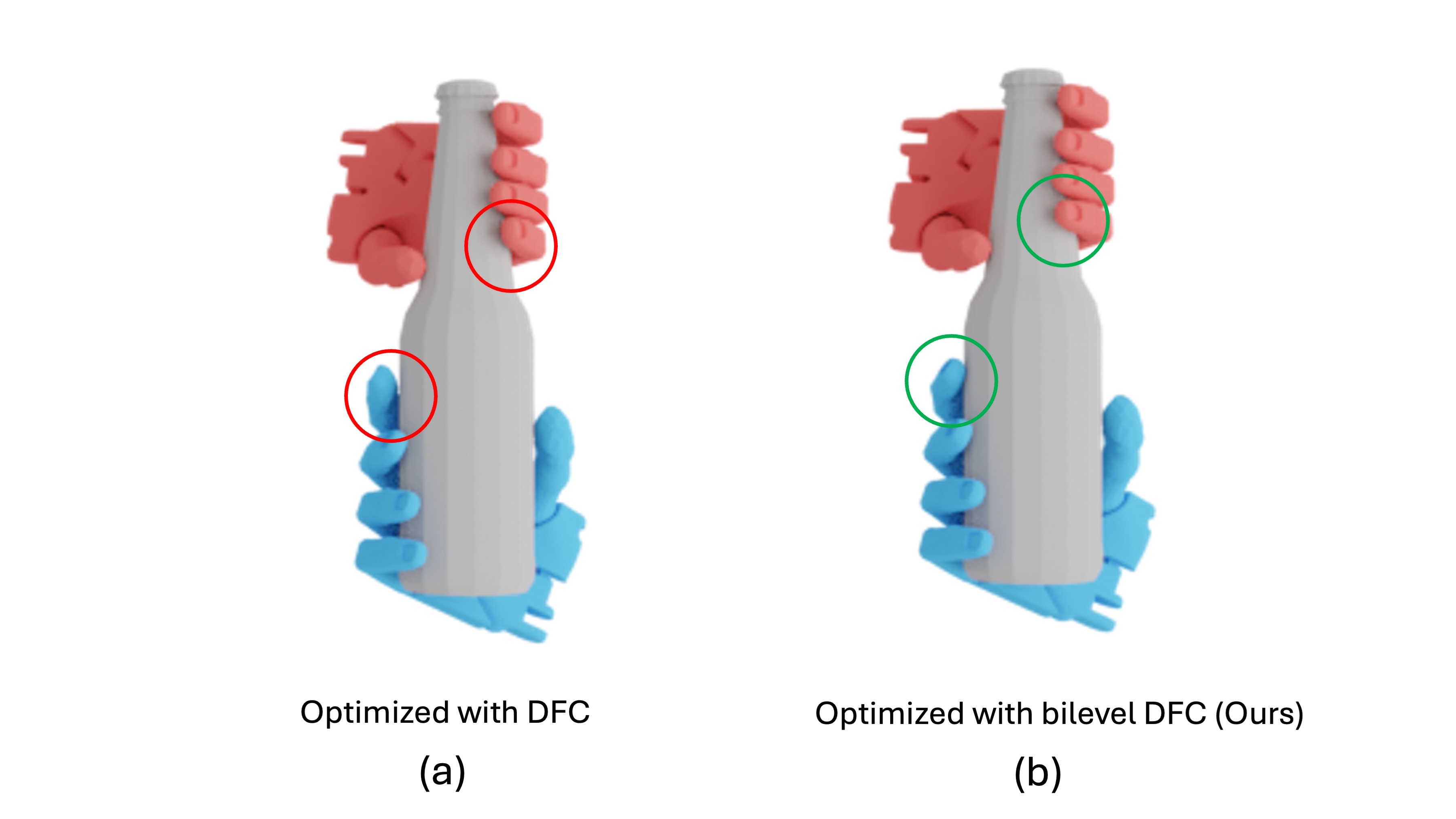}
    \vspace{-18pt}
    \caption{\textbf{Visualization of LP-based DFC vs DFC.} \textbf{Left:} DFC generates artifacts such as tilted finger and drifted contacts, \textcolor{red}{highlighted in red}. \textbf{Right:} LP-based DFC generates more natural poses that are compliant with the geometry of the object, \textcolor{green}{highlighted in green}. The grasp poses are optimized from identical initializations and only differ by using DFC or LP-based DFC for optimization.}
    \label{fig:DFC}
\end{figure}

\subsection{LP-based DFC}
As analyzed in DexGraspNet 2.0~\cite{zhang2024dexgraspnet}, the assumption of equal-magnitude contact force in DFC limits the quality of synthesized grasp poses. To this end, we use linear programming that takes contact force magnitude into account. At each timestep, we first fix the hand pose and solve the optimal contact forces applied to each contact position, such that the net wrench applied to the object is minimized:

\begin{equation}
    \begin{aligned}
        P = &\min\limits_{\textbf{f}} \Vert G(\textbf{f} \odot c)\Vert_2\\
          s.t. & \max\limits_{i} (\textbf{f})_i = 1\\
               & (\textbf{f})_i \geq 0, i=1,2,...,n
    \end{aligned}
\end{equation}

Then we use the net wrench $P$ and corresponding contact forces \textbf{f} as hyperparameters to rescale the DFC energy when the current pose is already stable. Specifically, we check whether the total wrench is small enough and each contact point applies non-trivial force. If the current pose is not stable, we optimize with the original DFC to search for stable poses efficiently. We empirically find that grasps synthesized with LP-based DFC are more natural than those using DFC.

\begin{equation}
    E_{FC}=\left\{\begin{aligned}
        &\Vert G(\textbf{f} \odot c)\Vert_2, &\text{if  }P < \tau_{FC} \text{and } \min\limits_{i}(\textbf{f})_i \geq \tau_{f}\\
        &\Vert Gc\Vert_2, &\text{otherwise\quad\quad\quad\quad\quad\quad\quad\quad}
    \end{aligned}\right.
\end{equation}

A qualitative comparison between LP-based DFC and the original DFC is visualized in Fig.~\ref{fig:DFC}. Starting from the same initialization pose, we optimize the grasp poses with DFC and LP-based DFC, respectively, with identical hyperparameters and all other energy functions for optimization unchanged. Grasp poses synthesized with DFC suffer from artifacts of tilted fingertips and broken contacts. Under the assumption that each contact point applies equal contact pose, the DFC energy is not minimized at a natural grasp pose of the thumb opposing the rest 4 fingers. Therefore, minimizing with respect to DFC encourages the hand to drift away from the natural pose in this scenario. On the other hand, the LP-based DFC takes contact magnitude into account and produces grasp poses more natural and compliant to the geometry of the object.

\subsection{Regularization Energy}
We implement several regularization energies to counteract undesirable behaviors in the process of gradient-based optimization.

First, we would like to avoid both hand-object penetration and hand self-penetration in the optimization process. To this end, we incorporate the built-in implementation of penetration energy (denoted $E_{pen}$) and self-penetration energy (denoted $E_{spen}$) in the CuRobo~\cite{sundaralingam2023curobo} robotics library.

Second, we would like to prevent hand joint angles from going outside the limits. We incorporate the built-in implementation of joint limit energy in CuRobo~\cite{sundaralingam2023curobo} as well.

Third, in order to avoid synthesizing dexterous grasps in twisted poses, we encourage the hand-object contact points to align with the front side of the hand meshes. We use a cosine-similarity energy function
\begin{equation}
    E_{dir} = \sum\limits_{i=0}^n (1 - c_i \cdot N_i)
\end{equation}
where n is the total number of contact points on the hand, $c_i$ is the i-th contact normal vector on the object surface pointing inwards into the object, and $N_i$ is the normal vector of the front surface of the i-th contact link, pointing outwards from the hand link mesh.

In summary, the complete regularization energy is formulated as

\begin{equation}
    E_{reg} = \omega_{limit}E_{limit}+\omega_{pen}E_{pen}+\omega_{spen}E_{spen}+\omega_{dir}E_{dir}
\end{equation}
\section{Implementation Details of Object Preprocessing}
\label{sec:object}
The object assets in our DexGraspNet 3.0 dataset are preprocessed and filtered from the entire Objaverse~\cite{deitke2023objaverse} dataset, which encompasses over 800k 3D assets covering diverse object categories.

We first filter the assets by querying GPT-4o~\cite{achiam2023gpt}. Following SoFar~\cite{sofar25}, we concatenate images rendered from six orthogonal views of each object, and prompt GPT-4o with true-or-false queries of the following five requirements:
\begin{itemize}
	\item Whether the object is a scene or a collection of disconnected objects.
    \item Whether the object has single white or grey colors.
    \item Whether the object has a ground/tabletop plane.
    \item Whether the 3D model is a refined, well-constructed mesh without defects.
    \item Whether the object represents a recognizable and meaningful 3D object.
\end{itemize}

The detailed prompt used is identical to Fig.15 in~\cite{sofar25}. After filtering out undesired objects, we process the rest objects by extracting .obj meshes using Trimesh~\cite{trimesh}, and then obtaining a convex-decomposed watertight mesh with ManifoldPlus~\cite{huang2020manifoldplus} and CoACD~\cite{wei2022approximate}(with threshold=0.4). Low-quality objects that fail watertight remeshing, as well as too complex objects that generate more than 200 convex pieces are discarded.

\begin{figure*}[t!]
    \centering
    \includegraphics[width=0.84\linewidth]{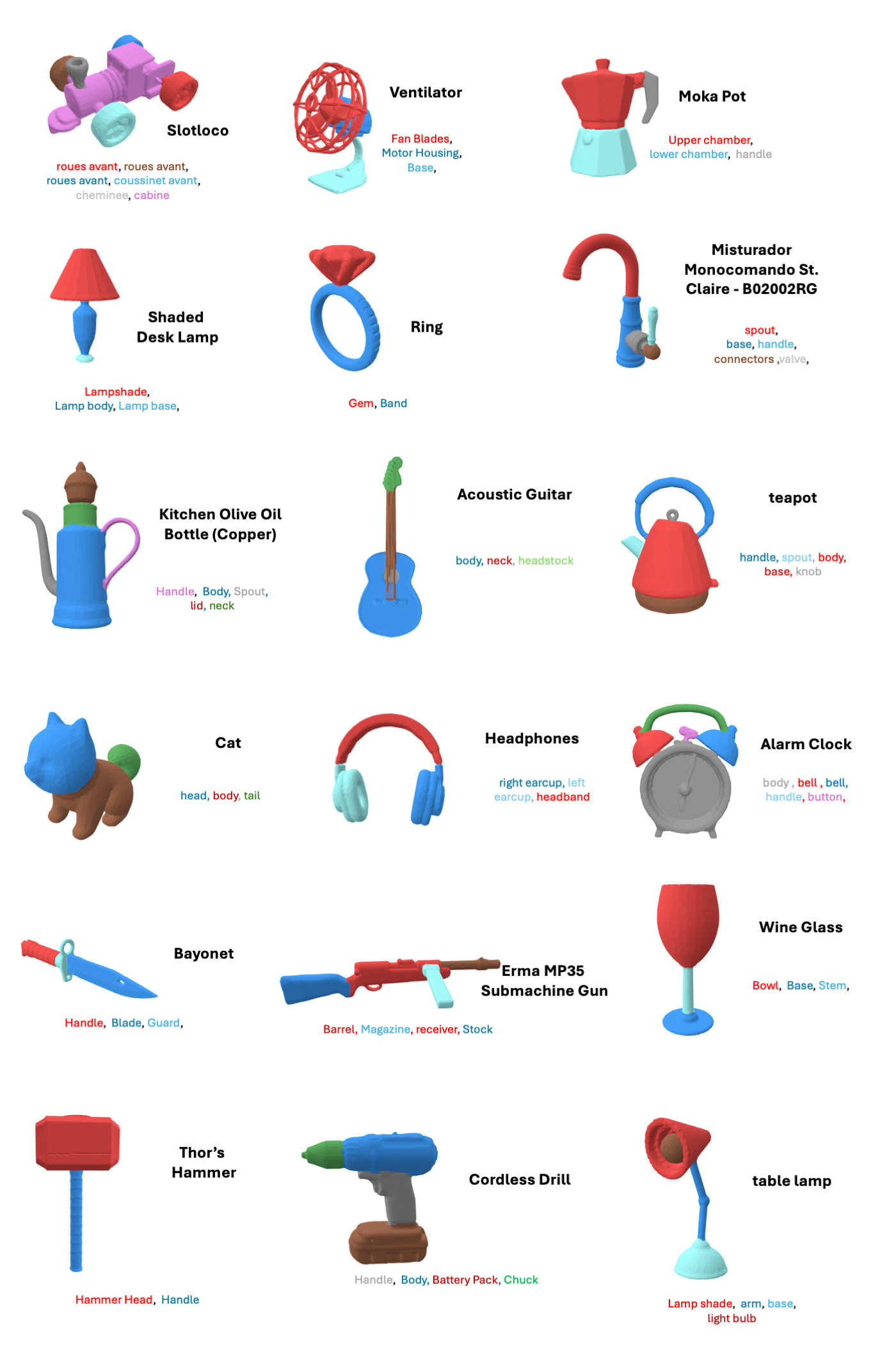}
    \vspace{-20pt}
    \caption{\textbf{Visualization of part segmentation and GPT generated part annotations.}}
    \label{fig:seg}
\end{figure*}

We use SAMesh~\cite{tang2024segment} to segment the remaining valid meshes into semantically meaningful parts. The segmented meshes are colored with segmentation masks and rendered from 6 orthogonal views. The multi-view images are then concatenated and fed into GPT-4o with Set-of-Marks prompting to obtain the semantic captions of each part. Visualization of sample segmented meshes together with the semantic part label is provided in Fig.~\ref{fig:seg}. The prompt used for part name labeling is given in Fig.~\ref{fig:prompt_partname}.

Finally, we ask GPT-4o to output a reasonable size of each object and rescale the objects accordingly. In order to make sure the semantically meaningful parts of each object are in a graspable size, we clip the diagonal length to the range [20, 50]cm. The prompt used to query object size is given in Fig.~\ref{fig:prompt_size}.

\section{Implementation Details of Hand Pose Initialization}
\label{sec:initialization}

This section provides implementation details of how we align the initial hand poses with the object part geometry. Given an object segmented into semantic parts, we sample a batch of 5000 initial hand poses around each part and perform gradient-based optimization to synthesize grasp poses aligned with the specified part. We first extract oriented bounding boxes of each part(Sec.~\ref{sec:bbox}). The oriented bounding boxes give geometric priors about the size, shape and orientation of object parts. The spatial relation among bounding boxes of different parts gives hints about the entire object's structure~\cite{sofar25,omnispatial25}. Leveraging these geometric cues, we align the initial position and orientation of the palm of the hands with object parts. Lastly, to synthesize diverse grasp poses, we jitter the initial palm poses and handcraft two sets of initial hand joint poses together with candidate contact points on the hand in Sec.~\ref{sec:splits}, resulting in the \textbf{Ours-Wrap} and \textbf{Ours-Pinch} splits of the DexGraspNet 3.0 dataset.

\begin{figure}[h!]
        \centering
        \includegraphics[width=0.85\linewidth]{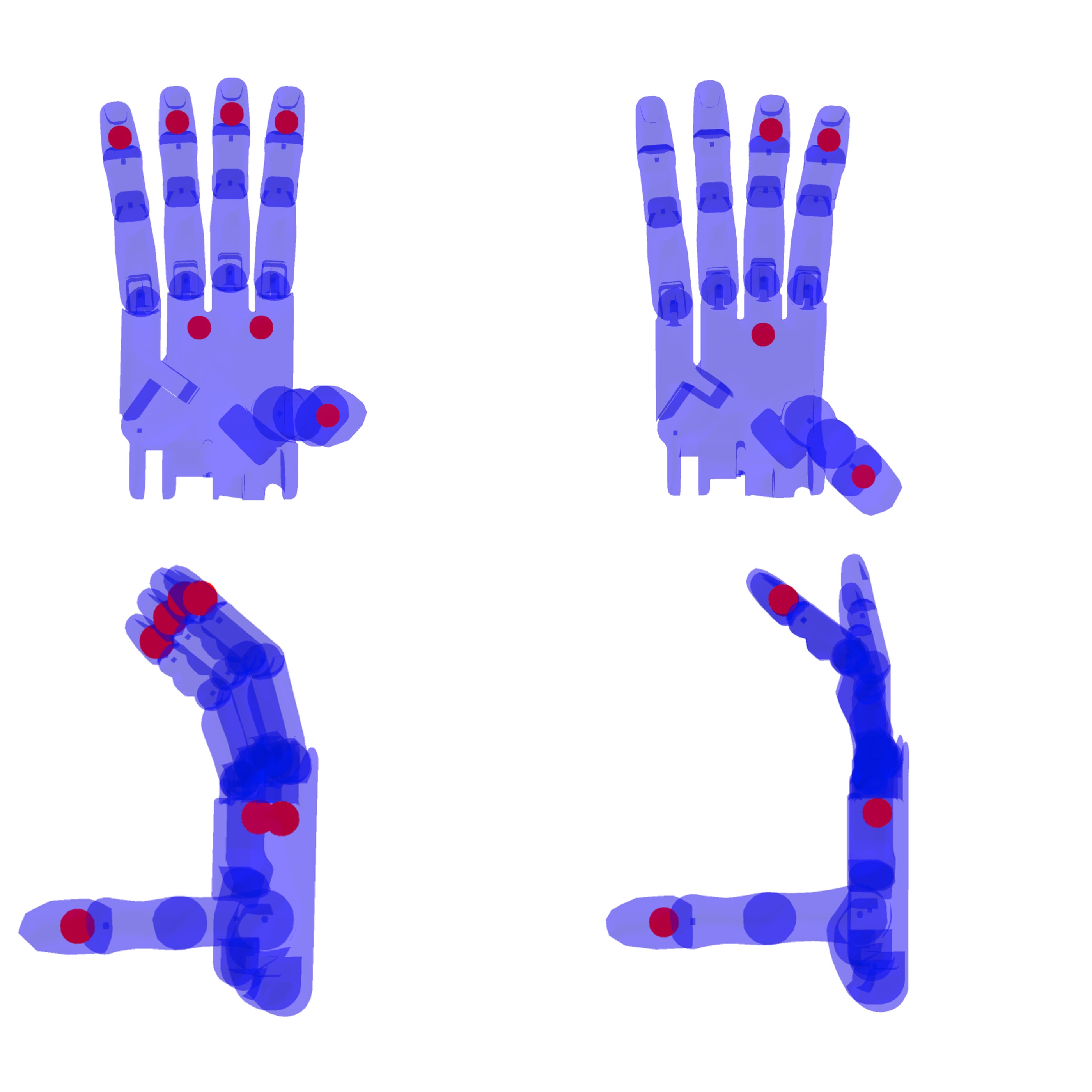}
        \vspace{-15pt}
        \captionof{figure}{\textbf{Visualization of initial hand joint positions and contact candidates.}}
        \label{fig:template}
        \vspace{-5pt}
\end{figure}

\begin{table}[h]
\centering
\resizebox{1.0\linewidth}{!}{
\begin{tabular}{l|cccc}
    \textbf{Split Name} & \textbf{Object} & \textbf{Grasps} & \textbf{Captions} & \textbf{Contact Candidates} \\
    \hline
    \textbf{Ours-Wrap} & 169k & 103M & 103M & 7 \\
    \hline
    \textbf{Ours-Pinch}& 139k & 67M & 67M & 4 \\
    \hline
    \textbf{Total} & 174k & 170M & 170M & - \\
\end{tabular}
}
\captionof{table}{\textbf{Statistics of DexGraspNet 3.0 dataset in splits.}}
\label{tab:splits}
\vspace{-5pt}
\end{table}

\subsection{Oriented Part Bounding Box Generation}
\label{sec:bbox}
We generate the bounding box of each object part in order to quantify the geometry and orientation of each part. In specific, we sample 256 directions using Fibonacci lattices~\cite{gonzalez2010measurement,fang2020graspnet} as the x-axis, and for each direction, compute the in-plane rotation for generating the tightest part bounding box following~\cite{he2023tracking}. We select the minimal volume box among these 256 candidates as the oriented bounding box of the object part.

We refer to the axis directions of the oriented part bounding box as \textbf{part principal directions}. The part principal directions build a local frame on the object part that indicates its orientation. 

\subsection{Aligning Initial Palm Poses With Part Geometry}
Depending on the direction, length and inter-part relation of part principal directions, we classify all object parts in the DexGraspNet 3.0 dataset into 4 categories, which are detailed as follows. For each category, we use specific strategies to sample grasp points on the surface of the part and align initial palm poses with the part's principal directions. Although our categorization is not rigorous nor exhaustive, we observe that most functional parts of objects are similar in shape and are designed to be grasped in a few specific styles. Moreover, as pointed out in~\cite{wang2023dexgraspnet}, the quality of synthesized grasp poses in gradient-based optimization is very sensitive with respect to initialization poses. By aligning hand initial poses with human-inspired heuristic rules, we inject a strong human-alignment prior into the optimization process, which greatly helps us to synthesize more natural and semantically distinguishable grasp poses. As visualized in Fig.~\ref{fig:alignment}, grasp poses synthesized with our meticulously designed initialization are more natural than those initialized in random poses.

We categorize the object parts into 4 categories: \textbf{lid-like},\textbf{disk-like},\textbf{L-shaped} and \textbf{shaft-like};

\textbf{Lid-like} part is typical for knobs, lids of containers and some types of handles. We categorize a part as a lid-like part if the following conditions are met:
\begin{itemize}
    \item At least 4 corners of the part bounding box are inside the bounding box of another part, which indicates the current part is ``embedded inside" this other part.
    \item Any 4 corners at the same end of the longest axis of the part bounding box should not be inside a single part bounding box. This excludes the case that the part is a shaft.
\end{itemize}

\begin{figure}[h!]
    \centering
    \includegraphics[width=0.75\linewidth]{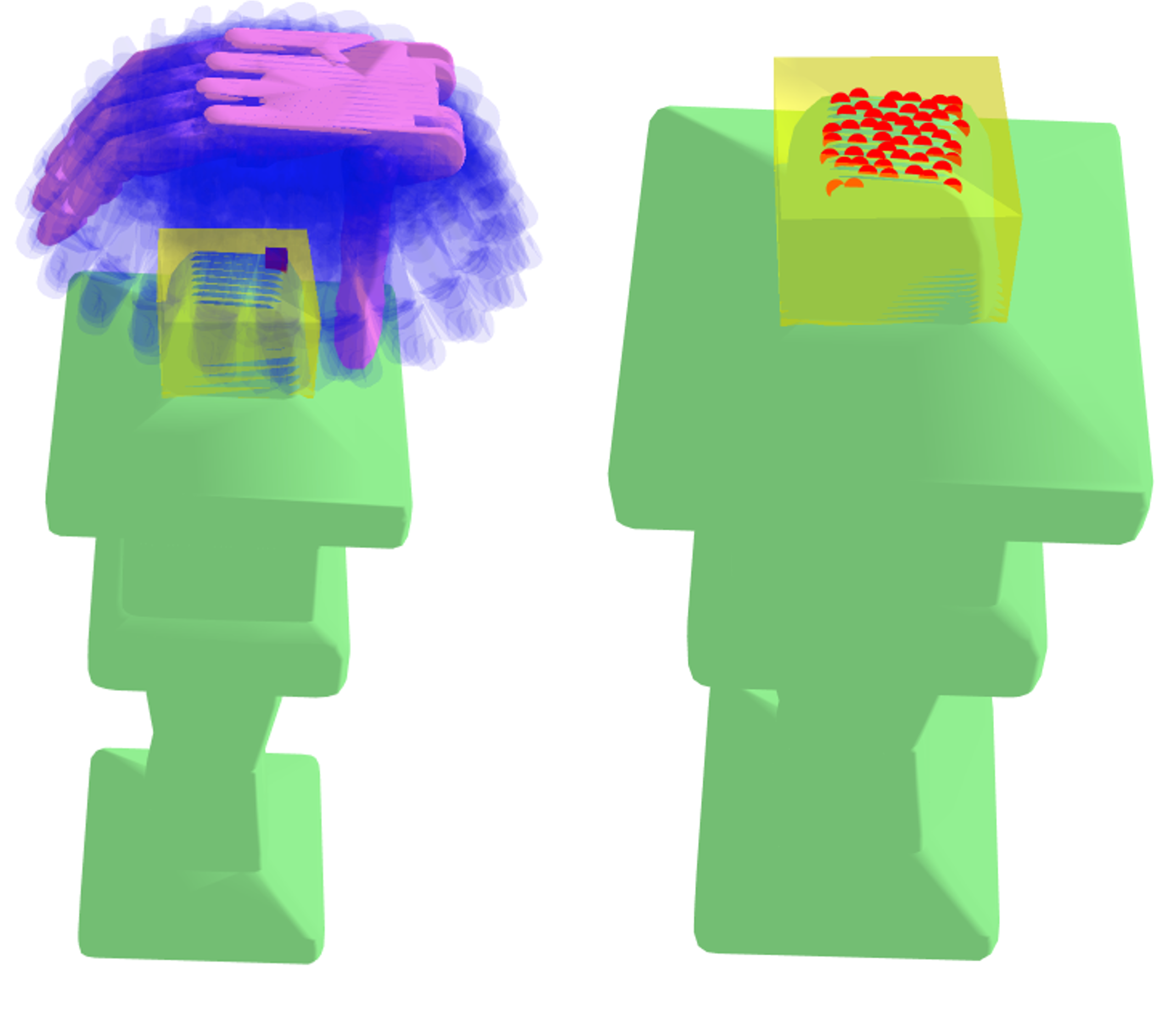}
    \vspace{-10pt}
    \caption{\textbf{Lid-like part}}
    \label{fig:lid_like}
    \vspace{-5pt}
\end{figure}

For lid-like parts, we identify the part's principal direction as the bounding box axis that points outside of the other part it is embedded in. We sample grasp points on the surface area where the surface normal aligns with the principal part direction. We align the palm to retreat from each grasp point and be perpendicular to the part principal direction and jitter by rotating the hand along the part principal direction for 24 uniform angles. One example of initialization with lid-like parts is visualized in Fig.~\ref{fig:lid_like}.

\textbf{Disk-like} part is typical for the supporting surfaces of objects, such as base of a lamp, the seat of a chair, and the tabletop surfaces. We categorize a part as a disk-like part if the following conditions are met:
\begin{itemize}
    \item The second-shortest axis is at least 3 times longer than the shortest axis, which indicates the part is flat.
    \item None of the corners of the part's bounding box is inside another part, which indicates the current part is protruding.
\end{itemize}

\begin{figure}[h!]
    \centering
    \includegraphics[width=0.8\linewidth]{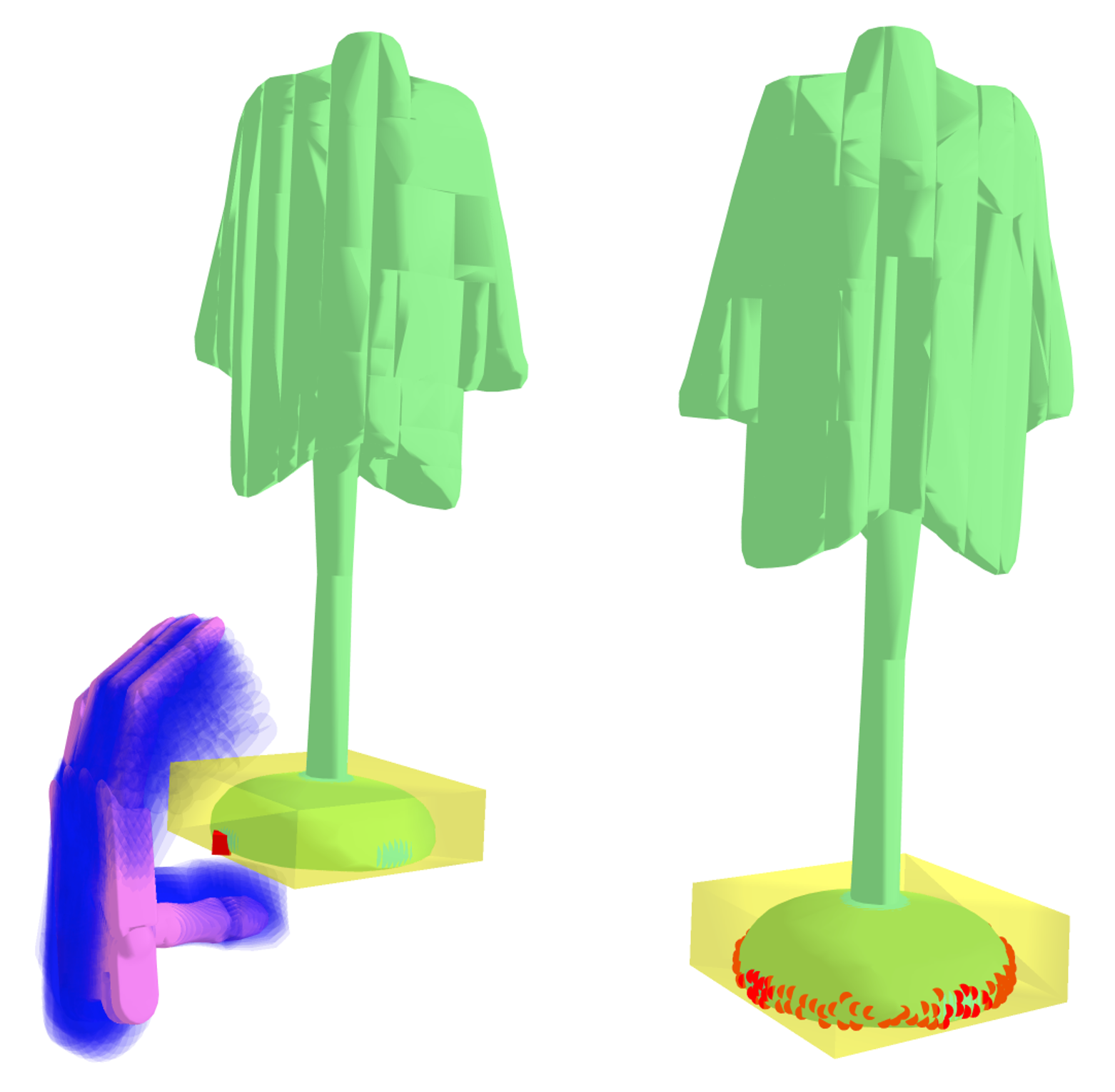}
    \vspace{-10pt}
    \caption{\textbf{Disk-like part}}
    \label{fig:disk_like}
\end{figure}

For disk-like parts, we identify the shortest axis of its bounding box that points into an adjacent part as the part principal direction. We sample grasp points on the surface area where the surface normal is perpendicular to the part principal direction. We align the palm to retreat from each grasp point with the front face of palm perpendicular to the surface normal of the grasp point and the up direction of palm align with the principal direction. We jitter the palm pose by slightly rotating it sideways, as shown in Fig.~\ref{fig:disk_like}.

\textbf{L-shaped} part is typical for thin object parts designed for human-object interaction, such as the headband of headphones, the seats of vehicles and the handle of kettles. We categorize a part as an L-shaped part if any corner section of the part's bounding box does not intersect with the mesh of the part. 

\begin{figure}[h!]
    \centering
    \includegraphics[width=0.9\linewidth]{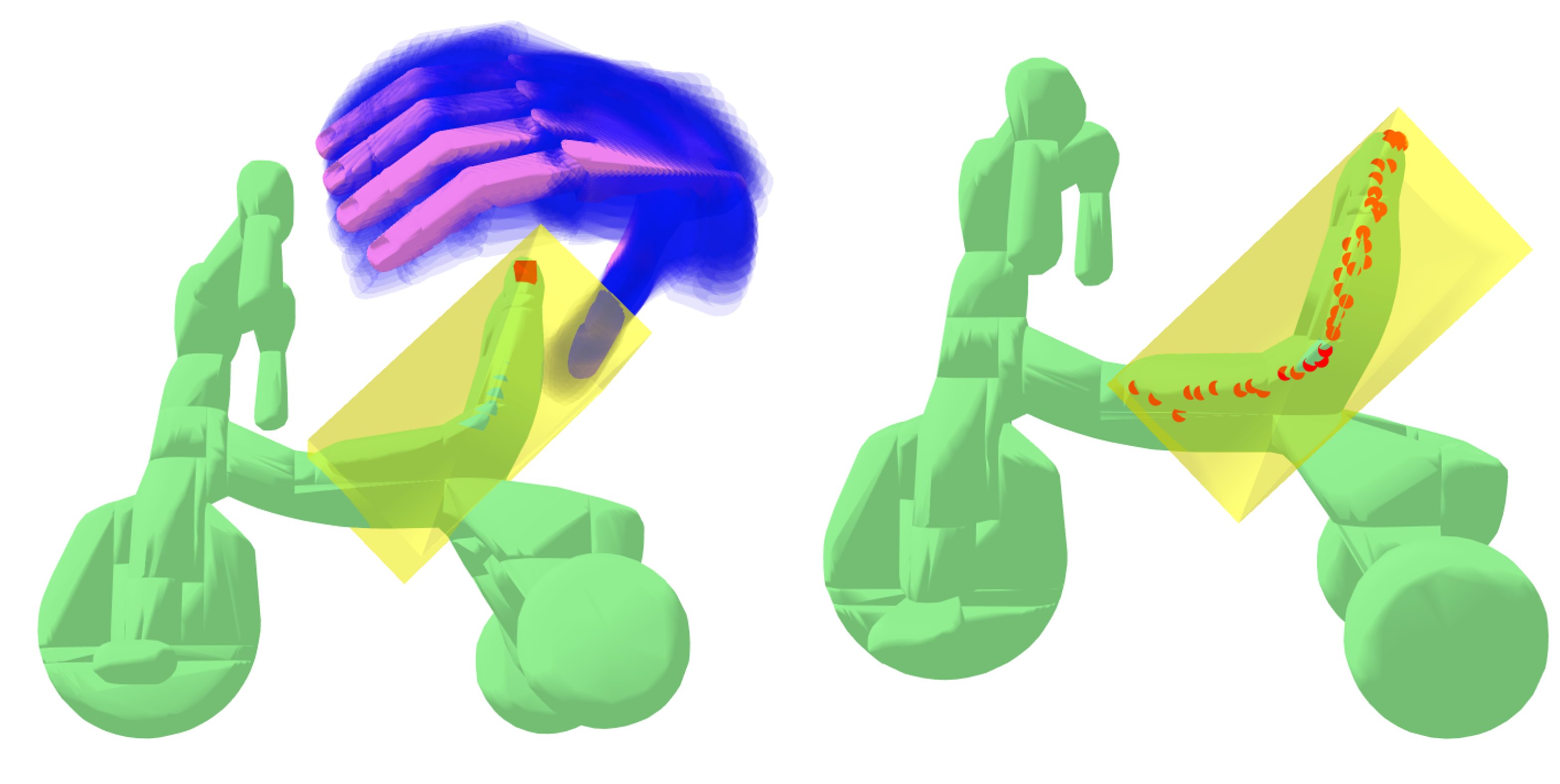}
    \vspace{-10pt}
    \caption{\textbf{L-shaped part}}
    \label{fig:L-shaped}
\end{figure}

For the L-shaped part, we identify the part principal direction as the direction pointing from the center of part bounding box to the middle of the bounding box edge corresponding to the missing corner section. We sample grasp points on the surface area where the surface normal is perpendicular to the principal part direction. We align the palm to retreat from each grasp point with the front face of the palm perpendicular to the surface normal of the grasp point and the up direction of the palm aligns with the direction pointing from the grasp point to the center of the part's bounding box. We jitter the palm pose by slightly rotating it sideways, as shown in Fig.~\ref{fig:L-shaped}.

\textbf{Shaft-like} part is typical for handles of tools and connecting bars of long objects such as lamps and barbells. Object parts that are not categorized into any of other types are categorized as shaft-like parts by default, because we emperically find the initialization strategy for shaft-like part is relatively more robust to uncommon geometries.
\begin{figure}[h!]
    \centering
    \includegraphics[width=0.9\linewidth]{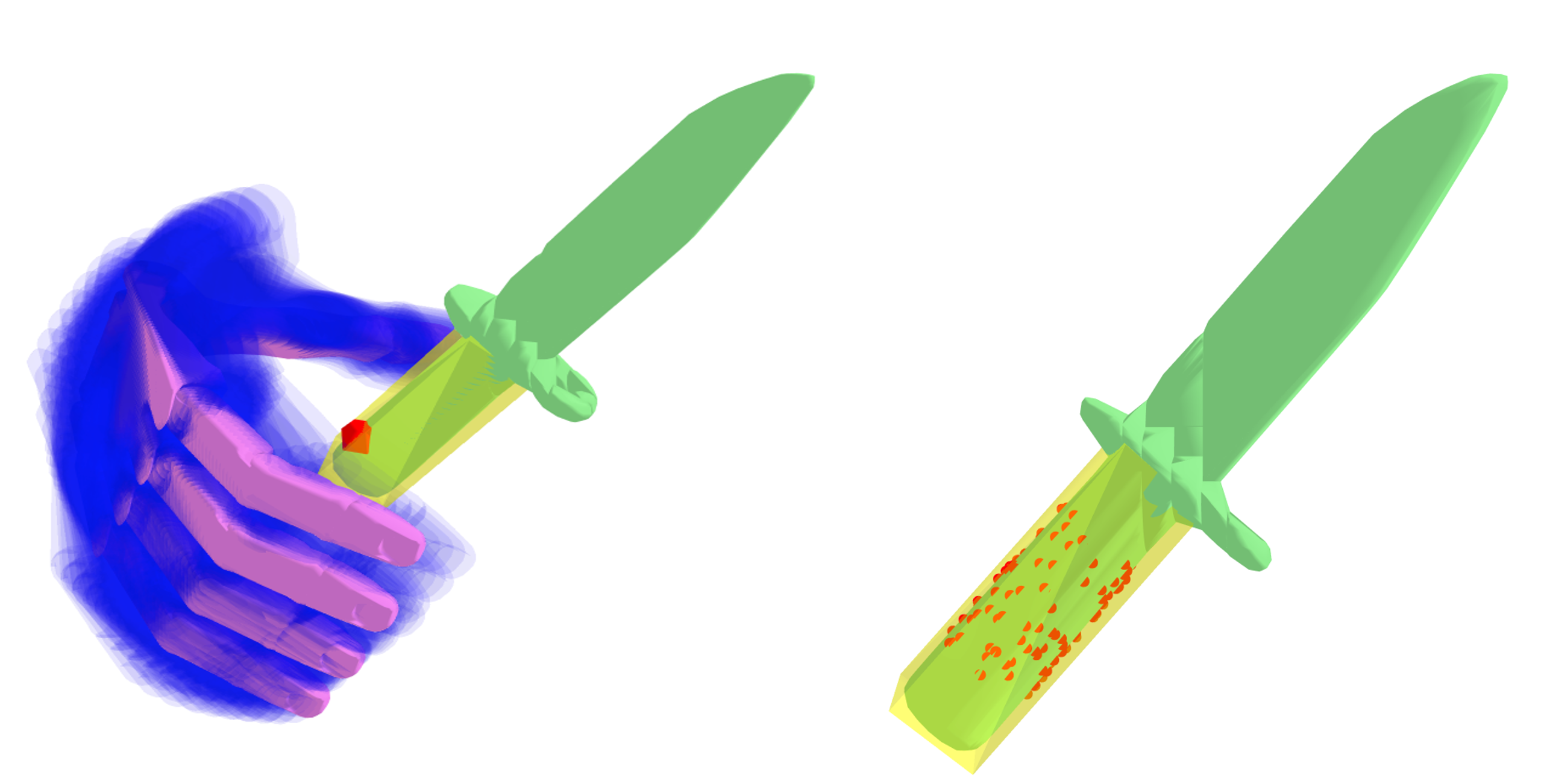}
    \vspace{-5pt}
    \caption{\textbf{Shaft-like part}}
    \label{fig:shaft_like}
\end{figure}

For shaft-like part, we identify the part principal direction as the longest axis of part bounding box aligned with the direction from part bounding box center to object bounding box center. We sample grasp points on the surface area where the surface normal is perpendicular to the part principal direction. We align the palm to retreat from each grasp point with the front face of palm perpendicular to the surface normal of the grasp point and the purlicue direction of palm aligned with the part principal direction. We jitter the palm pose by slightly rotating it sideways, as shown in Fig.~\ref{fig:shaft_like}.

\subsection{Hand Joint Pose and Collection of Contact Points}
\label{sec:splits}

Following~\cite{zhang2024dexgraspnet}, we handcraft initial joint angles and contact candidates of the hand, as visualized in Fig.3. Contact candidates refer to the labeled points on the hand mesh, with which we calculate the distance energy $E_{dis}$ and encourage these points to make contact with the object. In \textbf{Ours-Wrap} split, we set contact candidates on all 5 fingertips as well as on the palm. Grasp poses synthesized in In \textbf{Ours-Wrap} split usually wrap the object part in the hand. In In \textbf{Ours-Pinch} split we set contact candidates on the thumb, forefinger, middlefinger and palm. Grasp poses synthesized in \textbf{Ours-Pinch} split are dominantly pinch grasps, which are more suitable for grasping tiny object parts. For each part on each object, we generate a batch of 5000 initialization poses for each of \textbf{Ours-Wrap} and \textbf{Ours-Pinch} split, respectively for the optimization process.

\section{Camera Views Used to Render the Tabletop Dataset}
\begin{figure}[h]
    \centering
    \includegraphics[width=\linewidth]{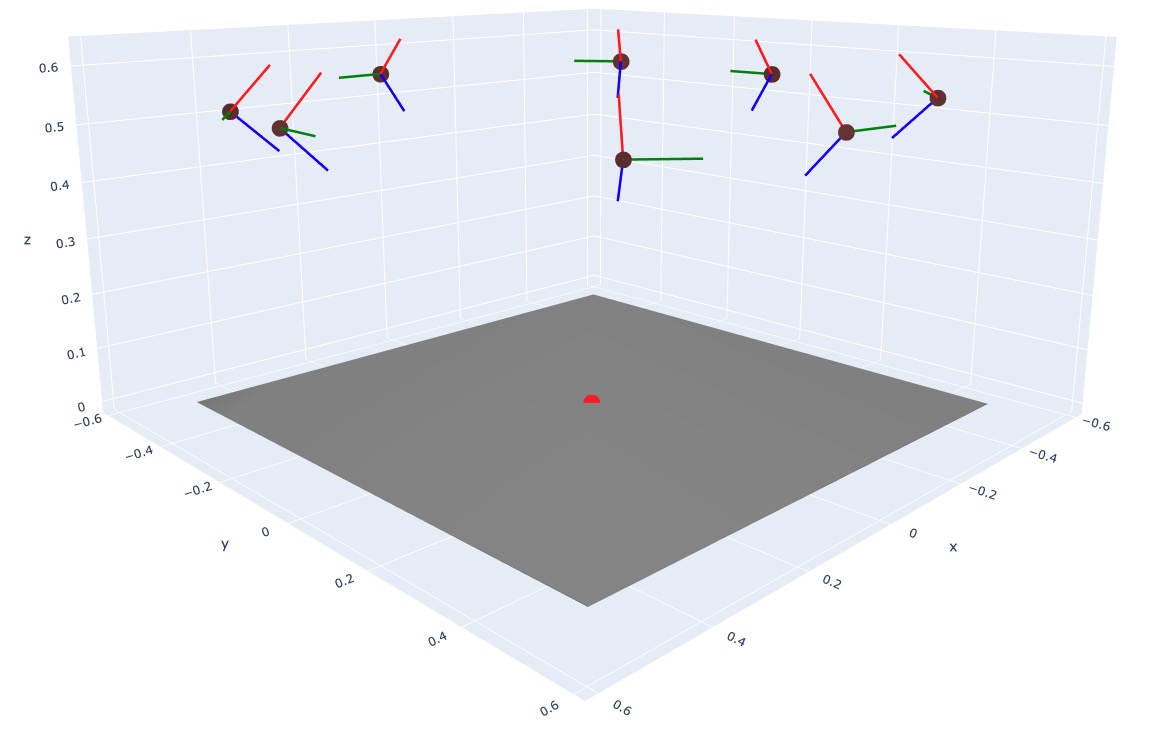}
    \vspace{-15pt}
    \caption{\textbf{Visualization of camera poses used to render the tabletop dataset}. We place the object on the table with translation in the x and y direction zeroed. The red dot represents the place where the object is placed. We sample camera poses evenly in the circle 80cm away from the table frame center and look down with 45 degrees in the yaw, looking at the origin. The camera frame (\textcolor{red}{+x},\textcolor{green}{+y},\textcolor{blue}{+z}) of each camera pose is visualized.}
    \label{fig:camera}
\end{figure}

\begin{figure*}[h]
    \centering
    \includegraphics[width=\linewidth]{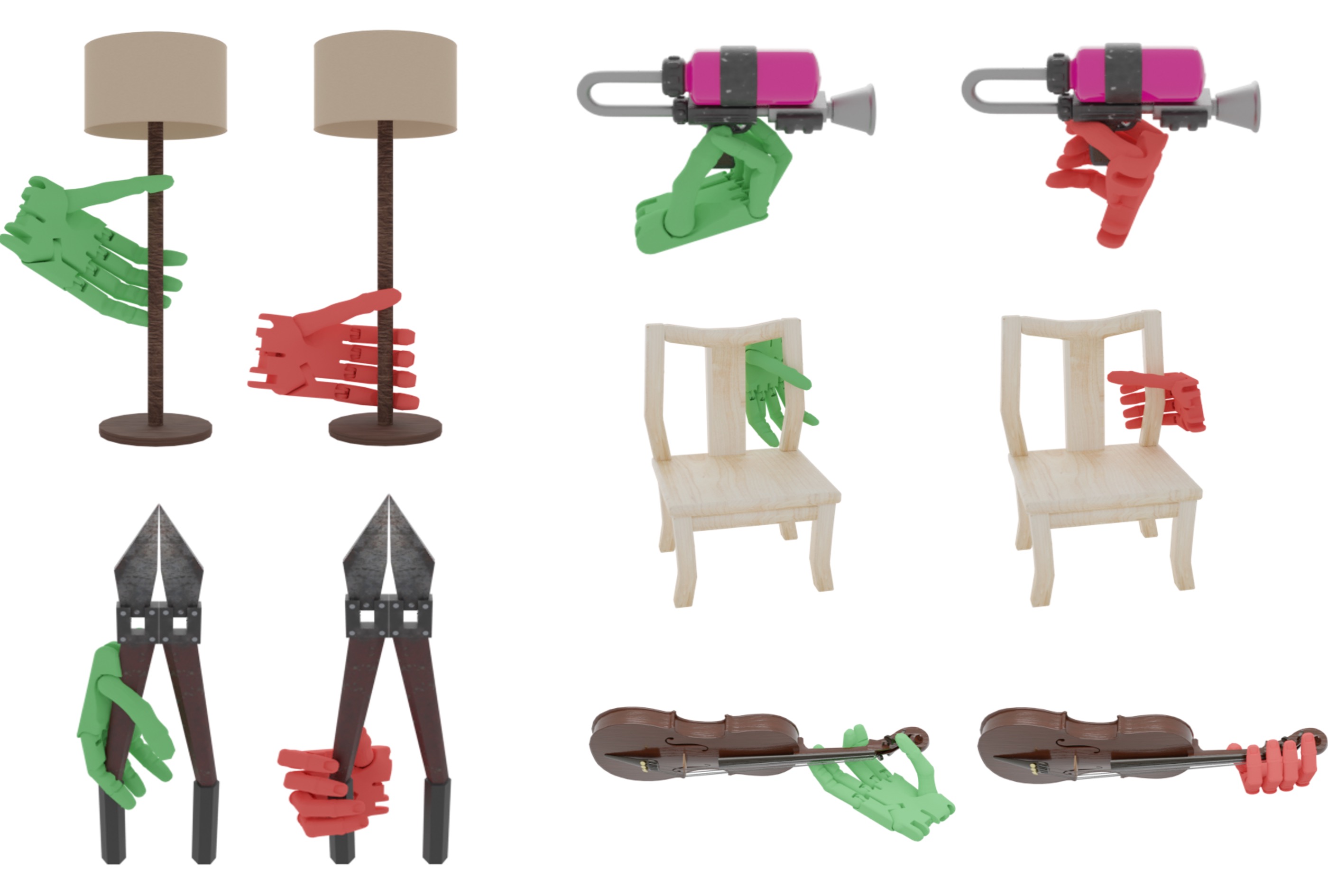}
    \caption{\textbf{Comparison between grasp poses synthesized with random initialization \textcolor{green}{colored in green} v.s. our part-aligned initialization \textcolor{red}{colored in red}}. The part-aligned initialization injects a strong prior about how a natural grasp is likely to be posed. The resulting optimized grasp poses with part-aligned initialization are more natural and semantically distinguishable, facilitating the instruction following training of VLG models.}
    \label{fig:alignment}
\end{figure*}

\begin{figure*}[h]
    \centering
    \vspace{-20px}
    \includegraphics[width=\linewidth]{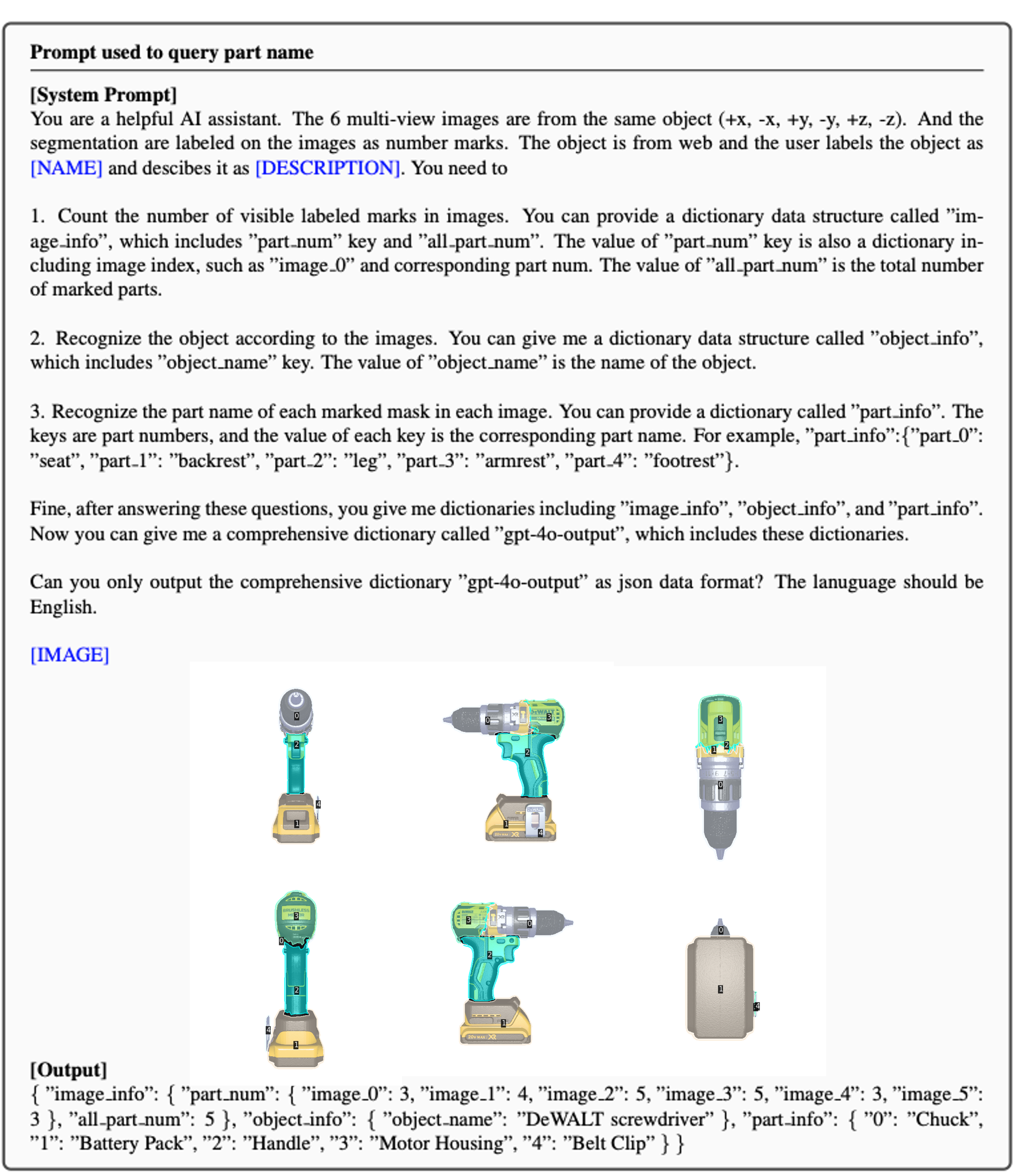}
    \caption{\textbf{Prompt used to query for part names of objects with GPT-4o.} We use Set-of-Marks~\cite{yang2023set} with images rendered from six views. \textcolor{blue}{[NAME]} and \textcolor{blue}{[DESCRIPTION]} are the captions given in the OpenShape~\cite{liu2023openshape} dataset.}
    \label{fig:prompt_partname}
\end{figure*}

\begin{figure*}[h]
    \centering
    \vspace{-20px}
    \includegraphics[width=\linewidth]{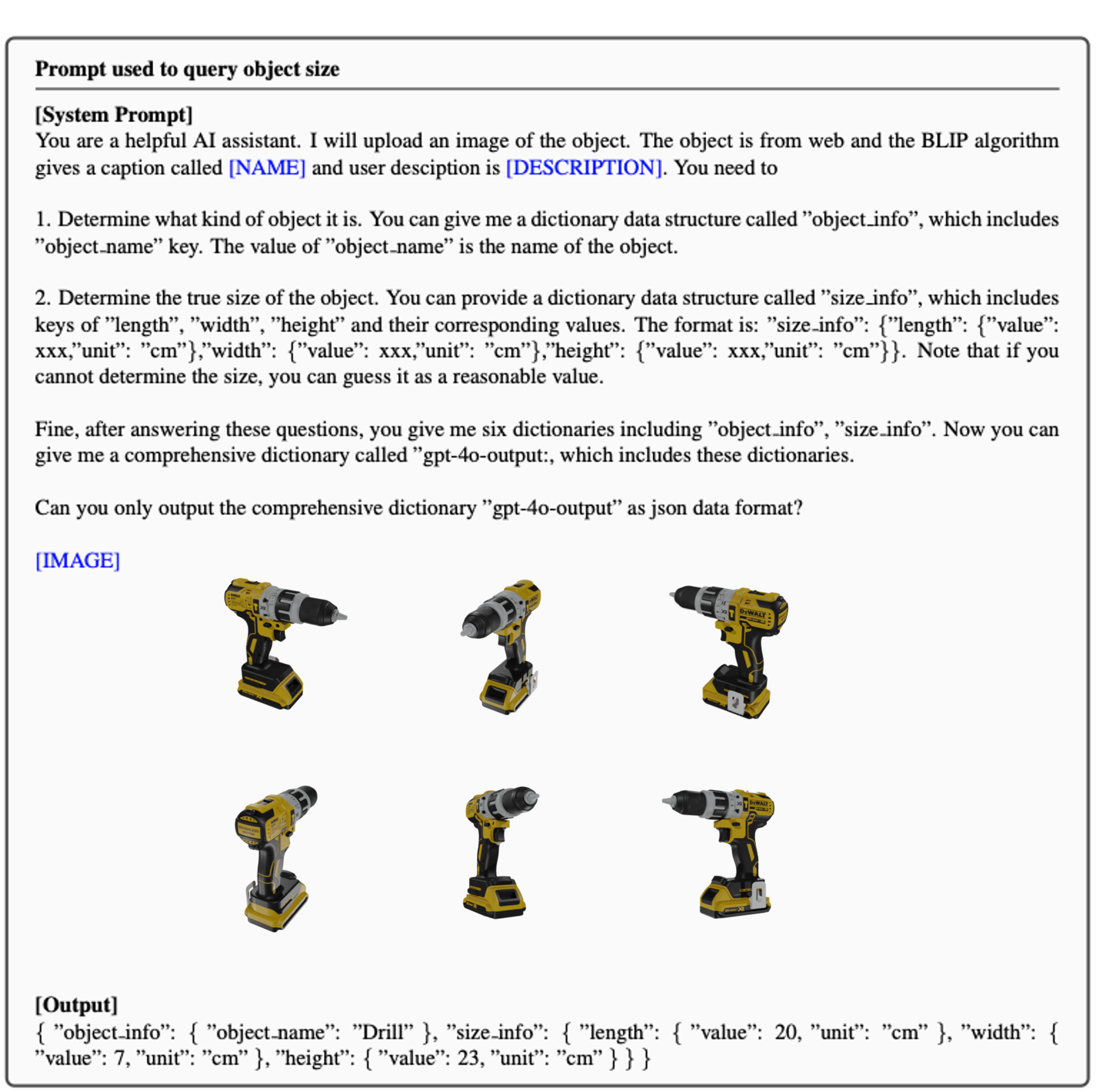}
    \caption{\textbf{Prompt used to query for object sizes with GPT-4o.}\textcolor{blue}{[NAME]} and \textcolor{blue}{[DESCRIPTION]} are the captions given in the OpenShape~\cite{liu2023openshape} dataset.}
    \label{fig:prompt_size}
\end{figure*}

We visualize the camera poses used to render the DexGraspNet 3.0 tabletop dataset in Fig.~\ref{fig:camera}. We place the object to grasp onto the table with stable poses collected from simulation, and manually set the translation in the x and y direction to zero. We evenly sample camera poses in the circle 80cm away from the table frame center and look down at 45 degrees in the yaw, looking at the origin.

\section{More Experiments}
\label{sec:more_experiments}

\subsection{Grasp Quality of DexGraspNet 3.0 Dataset}
Grasp poses in the DexGraspNet 3.0 dataset have less severe penetration and self-penetration, because we performed simulation-based penetration checking and filtered the undesirable poses. The poses in DexgraspNet 3.0 also demonstrate comparable stability quality compared against existing dexterous grasp datasets, with comparable Q1 statistics. Different from existing datasets, DexGraspNet 3.0 addresses the task of part-aligned grasp synthesis that emphasizes semantic alignment. The Q1 stability metric of DexGraspNet 3.0 is not expected to be higher than that in power grasp datasets, in which stability is the only goal of data synthesis.

\begin{table}[h]
    \centering
    \small
    \begin{tabular}{l|cccc}
    Method & \textbf{Scale}$\uparrow$ &  \textbf{Pen}$\downarrow$ & 
    \textbf{SPen}$\downarrow$ & \textbf{Q1}$\uparrow$\\
    \hline
    DexGraspNet~\cite{wang2023dexgraspnet}& 1.32M & 13.5& 0.93 & 0.114\\
    \hline
    DexGYS~\cite{wei2024grasp}& 50k & 3.32& - & 0.074\\
    SemGrasp~\cite{li2025semgrasp}& 50k & 1.1 & - & -\\
    Multi-GraspLLM\cite{li2024multi} & 120k & 7.1 & - & 0.091\\
    \hline
    \textbf{Ours-Wrap}   & 103M & 1.75 & 0.19 & 0.085\\
    \textbf{Ours-Pinch}   & 67M & 1.42 & 0.22& 0.067\\
    \end{tabular}
    \vspace{-5pt}
    \caption{\textbf{Evaluation of the dataset.} Pen-d and SPen-d are in mm.}
    \label{tab:metrics}
\end{table}

\subsection{More Ablation Experiments}
As shown in Table~\ref{tab:abl_style}, we also conduct more ablation experiments on grasp style.  
We find that the wrap grasp consistently outperforms the pinch grasp in both success rate and part graspability across all datasets, indicating greater stability and effectiveness in interacting with object parts. While the pinch grasp offers more flexibility and maneuverability, it struggles with stability, leading to lower performance. Both grasp styles see a drop in performance on unseen objects, highlighting challenges in generalization. This suggests a trade-off between stability and precision, where the wrap grasp is better suited for secure, robust grasps, while the pinch grasp may be preferable for delicate or precision-based tasks.

As shown in Table~\ref{tab:abl_diffusion}, the flow-matching paradigm is much better than traditional DDPM and DDIM diffusion-based action denoise modules. The simple flow-matching denoise module facilitates easier learning of pose generation.
\begin{table}[h]
    \centering
    \resizebox{\columnwidth}{!}{\begin{tabular}{l|cc|cc|cc}
    \toprule
    \multirow{2}{*}{Data}  &
    \multicolumn{2}{c|}{LVIS-Seen} & \multicolumn{2}{c|}{Unseen} & \multicolumn{2}{c}{SamPart3D}\\
    &  \textbf{Suc}$\uparrow$ & \textbf{PGA}$\uparrow$ & \textbf{Suc}$\uparrow$ & \textbf{PGA}$\uparrow$& \textbf{Suc}$\uparrow$ & \textbf{PGA}$\uparrow$\\
    \midrule
   Wrap grasp &87.7 &62.1 &79.1 &36.6 &76.3 &52.0\\
   Pinch grasp &71.8 &20.2 &54.8 &15.2 &50.6 &21.3\\
    \bottomrule
    \end{tabular}
    }
    \caption{\textbf{Ablation study for grasp style }.}
    \label{tab:abl_style}
\end{table}
\begin{table}[h]
    \centering
    \resizebox{\columnwidth}{!}{\begin{tabular}{l|cc|cc|cc}
    \toprule
    \multirow{2}{*}{Method}  &
    \multicolumn{2}{c|}{LVIS-Seen} & \multicolumn{2}{c|}{Unseen} & \multicolumn{2}{c}{SamPart3D}\\
    &  \textbf{Suc}$\uparrow$ & \textbf{PGA}$\uparrow$ & \textbf{Suc}$\uparrow$ & \textbf{PGA}$\uparrow$& \textbf{Suc}$\uparrow$ & \textbf{PGA}$\uparrow$\\
    \midrule
   DDPM &51.9 &7.8 &34.1 &10.9 &40.7 &5.5\\
   DDIM &57.7 &12.5 &39.6&10.4 &35.2 &8.5\\
   \midrule
   FlowMatching &75.3 &39.1 &54.0 &18.3 &53.4 &27.0\\
    \bottomrule
    \end{tabular}
    }
    \vspace{-5pt}
    \caption{\textbf{Ablation study for denoising paradigms used in pose prediction}.}
    \label{tab:abl_diffusion}
    \vspace{-5pt}
\end{table}

\clearpage

\clearpage

\end{document}